\documentclass[11pt, twocolumn]{article}
\usepackage[a4paper, margin=1.5cm]{geometry}
\usepackage{chemformula}
\usepackage[T1]{fontenc}
\usepackage{amsmath}
\usepackage{amssymb}
\usepackage{amsfonts}
\usepackage{graphicx}
\usepackage[labelfont=bf]{caption}
\usepackage{bbm}
\usepackage{booktabs}
\usepackage{multirow}
\usepackage{longtable}
\usepackage{makecell}
\usepackage{adjustbox}
\usepackage{authblk}
\usepackage{placeins}
\usepackage[superscript, nomove]{cite}
\usepackage{url}
\usepackage[many]{tcolorbox}
\usepackage{soul}
\usepackage{wasysym}
\usepackage{CJKutf8}
\usepackage{hyperref} 
\usepackage{tipa}
\usepackage[version=4]{mhchem}
\usepackage{booktabs} 
\usepackage{siunitx}  
\usepackage{adjustbox} 
\usepackage[ruled]{algorithm2e}
\newcommand{\mycommentstyle}[1]{\color{gray}{\small #1}}
\SetKwComment{Comment}{\mycommentstyle{// }}{}

\newtcolorbox{myminted}[2][]{colframe=black!60!black, colback=black!5!white, coltitle=white, title=#2,#1, fonttitle=\large, fontupper=\small}

\usepackage{floatrow}
\usepackage[label font=bf,labelformat=simple]{subfig}
\usepackage{caption}
\newcommand{\reasoningmodel}{QFANG}

\title{A Scientific Reasoning Model for Organic Synthesis \\Procedure Generation}

\author{
Guoqing Liu$^{1 *}$,
Junren Li$^{1, 2 * \dagger}$,
Zihan Zhao$^{1, 3 * \dagger}$,
Eray Inanc$^{1}$,
Krzysztof Maziarz$^{1}$,
Jose Garrido Torres$^{1}$,
Victor Garcia Satorras$^{1}$,
Shoko Ueda$^{1}$,
Christopher M. Bishop$^{1}$,
Marwin Segler$^{1}$
\\
\quad
\\
$^{1}$Microsoft Research AI for Science;
$^{2}$Peking University;
$^{3}$Shanghai Jiao Tong University;\\
*Equal contribution;
${^\dagger}$During an internship at Microsoft Research AI for Science\\
Correspondence to: \texttt{\{guoqingliu, marwinsegler\}@microsoft.com}
}

\date{}

\begin{document}

\maketitle

\begin{abstract}
Solving computer-aided synthesis planning is essential for enabling fully automated, robot-assisted synthesis workflows and improving the efficiency of drug discovery.
A key challenge, however, is bridging the gap between computational route design and practical laboratory execution, particularly the accurate prediction of viable experimental procedures for each synthesis step.
In this work, we present \reasoningmodel
\footnote{\reasoningmodel{} is a moniker derived from the Chinese phrase \textit{Qianfang} (meaning “thousands of recipes”). One suggested pronunciation is \textit{Chien-fahng}.},
a scientific reasoning language model capable of generating precise, structured experimental procedures directly from reaction equations, with explicit chain-of-thought reasoning.
To develop \reasoningmodel{}, we curated a high-quality dataset comprising 905,990 chemical reactions paired with structured action sequences, 
extracted and processed from patent literature using large language models.
We introduce a Chemistry-Guided Reasoning (CGR) framework that produces chain-of-thought data grounded in chemical knowledge at scale.
The model subsequently undergoes supervised fine-tuning to elicit complex chemistry reasoning.
Finally, we apply Reinforcement Learning from Verifiable Rewards (RLVR) to further enhance procedural accuracy.
Experimental results demonstrate that \reasoningmodel{} outperforms advanced general-purpose reasoning models and nearest-neighbor retrieval baselines, measured by traditional NLP similarity metrics and a chemically-aware evaluator using an LLM-as-a-judge.
Moreover, \reasoningmodel{}  generalizes to certain out-of-domain reaction classes and adapts to variations in laboratory conditions and user‑specific constraints.
We believe that \reasoningmodel{}’s ability to generate high-quality synthesis procedures represents an important step toward bridging the gap between computational synthesis planning and fully automated laboratory synthesis.

\end{abstract}

\section*{Introduction}

Organic synthesis is the foundational engine of molecular innovation, enabling the creation of a wide range of life‑saving pharmaceuticals and other advanced functional molecules\cite{li2015synthesis, blakemore2018organicsynthesis}. 
While modern algorithms can design millions of novel molecules \textit{in silico}, the practical synthesis of these molecules remains a major bottleneck\cite{blakemore2018organicsynthesis, campos2019importance, stanley2023fakeuntilmake}. 
This process is typically resource‑intensive and depends heavily on the tacit knowledge and expert intuition accumulated by chemists over years of practice\cite{ley2015organicsynthesis}.

To accelerate this process, the field of Computer-Aided Synthesis Planning (CASP) has made significant strides in automating reaction design\cite{vleduts1963concerning, corey1969computer, szymkuc2016casp, davies2019digitization, strieth2020machine, tu2023predictive}. 
Advanced algorithms can now predict single-step reactions\cite{segler2017neuralsym, liu2017retrosynthetic, schwaller2019molecular, dai2019retrosynthesis, chen2021deep, zhong2022rsmiles, fang2023substructure, li2023retroranker, xie2023retrosynthesis, gainski2025diverse, maziarz2025chimera}, plan complex multi-step retrosynthetic routes\cite{segler2018planning, lin2020automatic, chen2020retrostar, xie2022retrograph, liu2023pdvn, li2024retrobleu, tripp2023retro, maziarz2025syntheseus},
and even suggest suitable reaction conditions\cite{gao2018condition, chen2024enhancingcondition, wang2025reacon, sun2025multicondition,shim2025recommending}. 
Yet, a critical gap remains between these high-level plans and their practical execution in the laboratory\cite{mehr2020automaticchem,
vaucher2020automated,
smiles2actions}.
Converting strategic plans into precise, step-by-step experimental procedures still requires substantial human effort to specify essential operational details, such as the reagent addition sequence, reaction durations, temperature gradients, work-up,
and purification methods. 
This challenge is further amplified in the context of robotic systems intended to automate and scale chemical reaction execution\cite{steiner2019organicautomatic, coley2019robotic}.
Early work to address this gap framed procedure generation as a sequence-to-sequence task, 
training Transformer- and BART-based models\cite{vaswani2017transformer, lewis2019bart} from scratch to translate chemical equations into ordered action steps
\cite{ smiles2actions,christofidellis2023unifying,vavskevivcius2024language}.
While pioneering, these methods were constrained by the representational capacity of early models,
often struggling to produce lengthy, coherent procedures or to capture the underlying chemical principles of reactions.

\begin{figure*}[t]
    \centering
    \includegraphics[width=\textwidth]{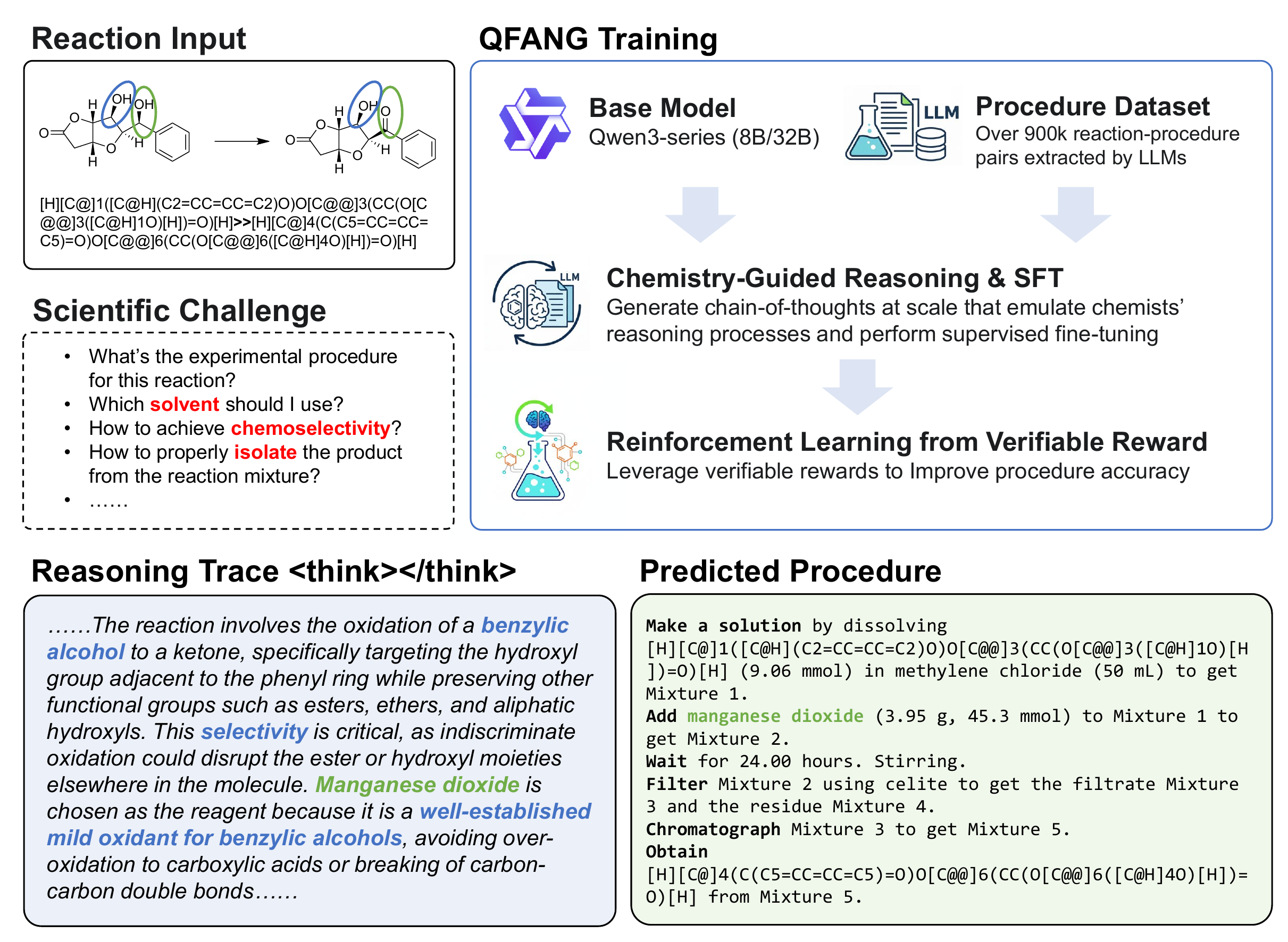}
    \caption{
    Overview of the inputs, training methodology, and outputs of \reasoningmodel{}.  
    The training process comprises three stages:
    (i) A large-scale experimental procedure dataset with over 900k reaction-procedure pairs is is assembled,
    (ii) A chemistry-guided reasoning (CGR) framework is applied to generate chemically grounded chain‑of‑thoughts at scale, followed by supervised fine‑tuning, and 
    (iii) A reinforcement learning stage that leverages verifiable rewards to further improve model predictions.
    During inference, \reasoningmodel{} takes a reaction equation as input, generates a high‑level reasoning trace capturing key decisions, and outputs a detailed, machine‑readable experimental procedure.
    }
    \label{fig:intro}
\end{figure*}

The recent advent of Large Language Models (LLMs) provides a promising avenue for this domain\cite{gpt4, jaech2024o1, anthropic2024sonnet35, comanici2025gemini, boiko2023autonomous, jablonka2024llmchem, m2024augmenting, zhao2025chemdfm, xia2025naturelm, bran2025chemical, mirza2025framework, zhang2025chemma, ether0, li2025ChemCoTBench, zhao2025chemdfmR, zhao2025molreasoner, wang2025chemR, zhao2025superchem}.
Trained on extensive corpora comprising chemical literature, reaction databases, and experimental protocols, LLMs internalize the statistical patterns that underlie established principles, precedent reactions, and methodological variations accumulated over decades of research.
Combined with their ability to interpret complex contexts and perform multi-step reasoning\cite{ouyang2022training,wei2022cot,guo2025deepseek},
LLMs are well‑suited for tasks like procedure generation.
Unlike conventional reaction condition prediction, which focuses on the core reaction stage, full procedure generation needs to specify the work-up and purification operations.
These operations require context-dependent chemical reasoning that current condition prediction models cannot provide\cite{matous2021workup, tzschucke2002workup}.

Initial efforts to harness LLMs have used few-shot, in-context learning with general-purpose models, guided by curated, action‑level datasets\cite{chen2025reactgpt}. 
However, in-context learning depends on analogical inference from limited examples, and thus struggles to develop a mechanistic understanding of chemistry. 
To build robust and capable systems, models should move beyond analogical imitation and undergo explicit training in chemical reasoning, enabling generalization beyond seen examples\cite{ether0,zhao2025chemdfmR, wang2025chemR}.

To address this gap, we developed \reasoningmodel{}, a scientific reasoning model post-trained on a large corpus of reaction–procedure pairs collected through an LLM-based automated annotation pipeline.
To elicit complex chemical reasoning, we propose Chemistry-Guided Reasoning (CGR), a two-stage framework that first programmatically constructs a \textit{factual skeleton} capturing the core logic of each chemical transformation, and subsequently expands it into an expert-style chain‑of‑thought using an LLM.
The model is supervised fine-tuned on the resulting CGR dataset to acquire chemical reasoning patterns.
To further improve model accuracy, we incorporate Reinforcement Learning from Verifiable Rewards (RLVR), which applies rule-based, step-wise reward functions to enforce chemical robustness.
A high-level overview of \reasoningmodel{} is shown in Figure~\ref{fig:intro}.
We instantiate the first version of \reasoningmodel{} on a dataset comprising 905,990 reaction-procedure pairs, while noting its straightforward scalability to larger datasets in the future.

To evaluate \reasoningmodel{}'s capabilities, we benchmarked it against advanced general-purpose reasoning models and nearest-neighbor retrieval baselines, including GPT-5 (High)\footnote{https://openai.com/index/introducing-gpt-5}, using established text-similarity metrics consistent with prior work.
Recognizing the limitations of traditional metrics, we further developed a chemically-aware evaluation framework that leverages GPT-5 as an expert judge.
On conventional metrics, \reasoningmodel{} 
achieves a BLEU-4 score of 61.3, surpassing the 54.4 score of a strong retrieval-augmented 3-shot GPT-5 baseline.
Under expert-judge evaluation, this margin increases further.
Additional analyses show that \reasoningmodel{} generalizes well to out-of-domain reactions, adapts procedural plans to chemist-specified constraints, and even corrects flawed procedures originating from its training data.
These results demonstrate that \reasoningmodel{} exhibits a deeper chemical understanding that extends beyond simple pattern matching.

\section*{Large-Scale Procedure Dataset Construction via LLM Annotation}

To develop a reasoning model capable of generating viable experimental procedures, it is essential to first define a fine‑grained and comprehensive action system\footnote{A standard set of laboratory operations, e.g., Make solution, Add,  Change temperature, Quench.}, and collect large‑scale, high‑quality, structured procedure datasets.

\subsection*{Action System Design}
For an experimental procedure to be precise, structured, and machine‑readable, it should be grounded in a well‑defined action system that comprehensively covers all operations involved in chemical experimentation, with detailed specifications for each operation
\cite{mehr2020automaticchem, vaucher2020automated, ai2024extracting, zhong2024actionie, liu2024reactxt, zhao2025literature,yuan2025uspto, machi2025actionsurvey, zhang2025chemactor, mendes2025automated}.
Rather than directly adopting the earlier action system proposed by Vaucher et al.~\cite{vaucher2020automated} in 2020, which has been employed in several subsequent works such as ActionIE~\cite{zhong2024actionie} and OpenExp~\cite{liu2024reactxt}, we employ an enhanced action system that is both more expressive and more comprehensive.

Our system defines 24 distinct actions, 
extending the set with operations such as \texttt{Change pressure}, \texttt{Change atmosphere}, \texttt{Sample}, \texttt{Irradiate}, \texttt{Chromatograph}, and \texttt{Distill}.
Each operation is supported by more detailed arguments and outputs, enabling finer granularity and richer semantics.
In addition, the system supports parallel operations across multiple mixtures, whereas OpenExp focuses on procedures for a single target mixture.
Detailed definitions of all 24 actions are provided in the Appendix.
To assess the validity of our action system, we compared procedures for identical reactions represented using our system and OpenExp's, evaluated with the OpenAI O3-high scoring metric (defined in the Appendix). As shown in Table~\ref{tab:action_system}, our action system achieves superior performance in substance accuracy, action coverage, and procedural order.

\begin{table}[t]
    \centering
    \begin{tabular}{lcc}
        \toprule
         O3-high Scoring & OpenExp & Ours \\
         \midrule
         Substance Accuracy $\uparrow$ & 7.80 & 9.46 \\
         Action  Coverage $\uparrow$  & 5.72 & 8.62 \\
         Order Correctness $\uparrow$ & 6.35 & 8.92 \\
         \midrule
         Overall Score  & 6.14 & 8.76 \\
         \bottomrule
    \end{tabular}
    \caption{Quality comparison between experimental synthesis procedures expressed using our action system and those of OpenExp.}
    \label{tab:action_system}
\end{table}

\subsection*{Annotation of Structured Experimental Procedures from Text Paragraphs}

\begin{figure*}[t]
    \centering
    \includegraphics[width=1.0\linewidth]{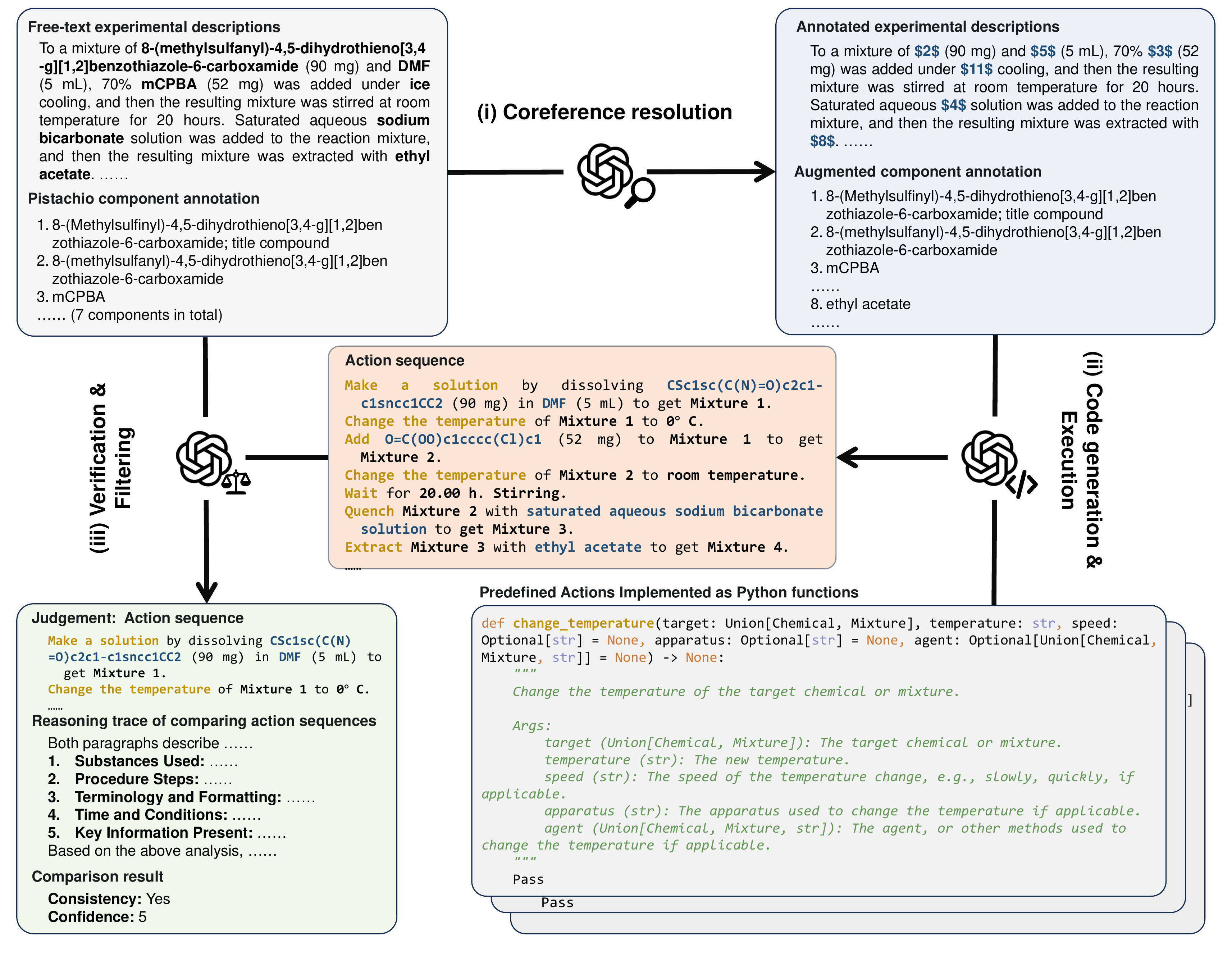}
    \caption{Overview of the procedure annotation pipeline, which consists of three main steps: 
    (i) Coreference resolution: GPT-4o annotates free-text experimental descriptions by replacing chemical entity mentions with their code names, assisted by component annotations from Pistachio.
    (ii) Code generation: GPT-4o translates the annotated experimental descriptions into structured, step-by-step procedures in Python.
    (iii) GPT-4o compares the generated procedures with the corresponding descriptions to validate annotation accuracy.
    }
    \label{fig:data_pipeline}
\end{figure*}

Constructing a high-quality, large-scale dataset of reaction–procedure pairs is essential for training reasoning models. Yet such data are typically expensive and labor‑intensive to obtain~\cite{lowe2012extraction, clark2015machines}.
Most existing chemical databases provide noisy, unstructured experimental descriptions mined from patents, rather than clean, step-by-step procedures.
To overcome this limitation and build a high‑quality, large‑scale dataset,
we leverage general-purpose LLMs.
While these models may lack deep chemical expertise, they are highly effective at extracting structured information and producing outputs in customized formats.
Specifically, we employ GPT‑4o\cite{hurst2024gpt4o} to transform free-text experimental descriptions from Pistachio\cite{pistachio}, one of the largest chemical reaction databases, into structured, step-by-step procedures following our defined action schema.
The overall annotation pipeline consists of three main steps as illustrated in Figure~\ref{fig:data_pipeline}.

\paragraph{(i) Coreference resolution.}
In a single textual description of an experiment, the same substance may be referred to by different names or pronouns.
To facilitate the subsequent translation into structured actions, we first instruct GPT‑4o to perform coreference resolution on each paragraph. Specifically, we provide GPT‑4o with the existing component annotations from the Pistachio database and ask it to replace all chemical entity mentions with their corresponding code names. In addition, GPT‑4o is allowed to augment the component list with newly identified molecules or materials when necessary. 
To ensure the accuracy of the coreference resolution process, we reconstruct the paragraphs and compute the edit distance between the original and reconstructed versions. Entries with edit distances exceeding a predefined threshold are discarded (Figure~\ref{fig:data_pipeline}(i)).

\paragraph{(ii) Code generation and execution.}
In this step, we implement each action in our action system as a python function, accompanied by supporting classes such as \texttt{Chemical}, \texttt{Mixture}, and \texttt{TimePeriod}. GPT‑4o is then instructed to translate each textual description into the corresponding Python code that invokes these actions.
Using code as an intermediate representation provides two key advantages: 1) it allows us to make use of advanced code generation abilities of LLMs, and 2) it enforces structural and type constraints, thereby ensuring that the resulting actions adhere to our pre-defined action specifications.
After LLM‑based annotation, the generated python code is executed and converted into the final action sequences, while samples with syntax errors or type‑check failures are removed (Figure~\ref{fig:data_pipeline}(ii)).

\paragraph{(iii) Verification and filtering.} In this step, we instruct GPT-4o to verify that the generated action sequences accurately express the same experimental procedure as their corresponding source paragraphs. GPT-4o is instructed to produce a reasoning trace before providing a judgment score for consistency (Yes, No, or Uncertain).
It then assigns a confidence score for the given judgment ranging from 0 to 5 (where 5 indicates the highest confidence, and 0 indicates the lowest). We retain all entries that pass the consistency verification with a confidence score above 3 and discard the rest (Figure~\ref{fig:data_pipeline}(iii)).
 
Detailed prompts used for all three steps are provided in the Appendix. 
Through the above action system design and automatic structured experimental procedure annotation pipeline, we obtain a high‑quality reaction‑procedure dataset from raw Pistachio, consisting of 905,990 chemical reactions paired with their corresponding structured action sequences. This dataset forms the foundation for the subsequent chemistry‑guided chain‑of‑thought data generation, supervised fine‑tuning, and reinforcement learning with verifiable rewards.

\section*{Chemistry-Guided Reasoning and Supervised Fine-Tuning}
While general-purpose LLMs have showed advanced reasoning in complex tasks such as mathematics and coding, their performance in specialized scientific domains, such as organic chemistry, remains a significant challenge \cite{mirza2025framework,li2025ChemCoTBench,zhao2025superchem}.
When tasked with generating a precise, in‑depth chain‑of‑thought (CoT) to derive suitable synthesis procedures, LLMs often generate reasoning that is verbose, factually inaccurate (hallucinated), or driven by superficial textual patterns rather than fundamental chemical principles\cite{llm_hallucinate, liu2025fgbench, li2025ChemCoTBench}. 
In the Appendix, we provide raw reasoning traces from general-purpose LLMs (Qwen3-Max\cite{qwen3max} and Phi-4-reasoning\cite{abdin2025phi})\footnote{We chose these open‑weight models because they allow access to their raw reasoning process.}
prompted with the example reaction shown in Figure~\ref{fig:intro}, illustrating the model’s difficulty in parsing the chemical structure.

Training a model to reason like a chemist requires data with two key criteria. First, the CoT reasoning must be grounded in chemical facts, accurately identifying strategic challenges such as chemoselectivity. Second, to support large-scale model training, such high-quality reasoning must be generated at scale, covering hundreds of thousands of chemical reactions like those present 
in the Pistachio dataset. Manual curation by domain experts is both time-consuming and prone to stylistic inconsistency, making it impractical for large datasets.

To address this, we developed \textbf{Chemistry-Guided Reasoning (CGR)}, a hybrid two-stage framework for generating large-scale, high-quality CoT datasets. In the first stage, a deterministic, cheminformatics‑based programmatic analysis is applied to chemical reactions to systematically identify key bond changes, functional group conversions, and strategic considerations.
This ensures that the underlying chemical logic is both factually accurate and structurally consistent across the entire dataset. 
In the second stage, these curated chemical facts serve as structured guidance for an LLM to generate coherent, expert-style reasoning narratives.
By combining the precision of programmatic analysis with the expressive power of LLMs, CGR enables scalable construction of chemically rigorous and stylistically consistent CoT datasets for training \reasoningmodel{}.

\paragraph{Programmatic Generation of Factual Skeletons.}
The first stage of CGR programmatically generates a structured, cheminformatics‑based reasoning trace for each reaction–procedure pair. It starts with a detailed analysis of the chemical transformation. Using an atom‑mapping algorithm (e.g., LocalMapper\cite{localmapper}), we compare the atomic connectivity and bond types in the reactants and products to identify which atoms and bonds are directly involved in the reaction. This structural analysis gives a clear picture of the transformation at the molecular level.

Building on this, our algorithm iterates through a predefined library of 243 functional groups, cataloging their presence in both reactants and products. It categorizes each group as being transformed, newly formed, or unchanged. This automated annotation allows the system to programmatically flag key strategic challenges that a chemist would need to address. For instance, in a selective oxidation of a benzylic alcohol in the presence of an aliphatic one, the system generated a key insight: \textit{"The functional group `alcohol' is converted to the functional group `aldehyde', while the other functional group `alcohol' remains unchanged. It should be noted that the `alcohol' functional groups are selectively transformed in this reaction. The procedure should take care of the selectivity issue."}

Concurrently, the associated experimental procedure is deconstructed. The algorithm programmatically separates the \textit{reaction} phase from the \textit{workup} phase by identifying keywords such as \texttt{Quench}, \texttt{Extract}, or \texttt{Filter}. Within each phase, it identifies all chemical entities and assigns their roles such as \textit{reactant}, \textit{catalyst}, \textit{reagent}, or \textit{solvent} following the raw annotations \cite{lowe2012extraction}. The script also captures critical procedural details, including the order of addition of these components and environmental conditions like the use of a protective nitrogen atmosphere. High-level context, such as the formal reaction name (e.g., "Horner-Wadsworth-Emmons reaction"), is also appended when identifiable. This stage produces a structured collection of verified chemical facts.

\paragraph{LLM-based Enhancement for Expert-like Narratives.}
In the second stage, the factual skeleton generated in the first stage serves as grounding context for an LLM (e.g., Qwen3-235B-Thinking-2507\cite{qwen3}). 
The input prompt provides the chemical reaction query, the ground-truth experimental procedure, and the programmatically generated factual skeleton. The model is explicitly instructed to synthesize these discrete facts into a coherent, explanatory narrative, adopting the perspective of an expert chemist designing the experiment. The prompt directs the model to emphasize the causal links between the identified challenges and the procedural decisions made.

Revisiting the selective oxidation example, where the proper oxidant here is manganese dioxide(\ce{MnO2}), the LLM enhances the rule-based skeleton to produce a final, expert-like reasoning process:
\textit{"The reaction involves the oxidation of a benzylic alcohol to a ketone, specifically targeting the hydroxyl group adjacent to the phenyl ring while preserving other functional groups such as esters, ethers, and aliphatic hydroxyls. This selectivity is critical, as indiscriminate oxidation could disrupt the ester or hydroxyl moieties elsewhere in the molecule. Manganese dioxide is chosen as the reagent because it is a well-established mild oxidant for benzylic alcohols, avoiding over-oxidation to carboxylic acids or breaking of carbon-carbon double bonds."}

This two-stage methodology ensures that the final CoT is not only factually accurate but also mirrors the causal, step-by-step logic of an expert chemist.

\paragraph{Supervised Fine-Tuning.}
Following the CGR procedure, all the 905,990 reactions with structured action sequences were paired with their corresponding chemical reasoning traces. 
Using this dataset, we performed a time-based split,
reserving the most recent 10\% of the entries as a held-out test set.
Specifically, the training set contains reactions published in source patents from 1971 to July 2023, while the held‑out test set covers those published from July 2023 to June 2024.
This time‑based split was chosen over a random split because it more accurately reflects real-world deployment, where a model trained on historical data is expected to predict future, previously unseen chemical procedures.

The remaining 90\% of the entries constituted the initial training corpus. 
A substantial portion of this first version dataset is derived from patents which, despite their scale, are often noisy, exhibiting issues such as stoichiometric inconsistencies, incomplete workup descriptions, and procedures optimized for intellectual property protection rather than experimental reproducibility.
To curate a high-fidelity training dataset, we implemented an LLM-based filtering protocol.
Specifically, we employed the Qwen3-235B-Thinking-2507 model as an expert chemical evaluator to score each annotated procedure (on a scale of 0–10) across four critical axes:
(i) \textit{Reactant Completeness}, (ii) \textit{Workup and Purification Completeness}, (iii) \textit{Condition Completeness}, and (iv) \textit{Reaction Safety}.
Only procedures achieving an average score of 5.0 or higher were prioritized, with the additional strict constraint that no single axis score could fall below 3.0. 
The detailed prompt used for scoring is provided in the Appendix.

With this curated dataset,  we initiated model training via Supervised Fine-Tuning (SFT).
We selected the Qwen-3 family\cite{qwen3} as base models, fine-tuning both the 8-billion and 32-billion parameter versions to examine the effect of model scale on performance.
During the SFT stage, the models were trained to produce coherent reasoning chains followed by structured experimental procedures.
These resulting SFT models, denoted as \reasoningmodel{}-8B (SFT) and \reasoningmodel{}-32B (SFT), not only demonstrated strong initial performance but also served as the starting policy for the subsequent reinforcement learning stage.

\section*{Reinforcement Learning with Verifiable Rewards}

After SFT, we leverage Reinforcement Learning from Verifiable Rewards (RLVR) to further enhance the prediction accuracy of ~\reasoningmodel{}.
However, predicting experimental procedures poses unique challenges for reward design.
In particular, exact matches between predicted and ground-truth procedures are difficult to achieve, and conventional text-similarity metrics like BLEU fail to capture chemical plausibility. As a result, commonly used outcome‑based reward functions in RLVR are not very suitable for this task.

To address this, we design a verifiable, step-wise reward function to guide the RL training.
For simplicity of verification, we assume that the predicted action sequence must strictly follow the ground-truth sequence.
Thus, synonymous but equivalent actions are treated as incorrect. This assumption enables step-by-step evaluation and the assignment of dense rewards. Specifically, the accuracy reward for each action at step $t$ consists of three parts.

\begin{description}
    \item[Format reward $R^t_{\mathrm{format}}$.]
    It checks whether the predicted action follows the correct format of any defined operation.
    If the action fails the format check, a negative reward of –1 is assigned, and no further rewards are computed for that step. If the format is correct, the reward for this component is zero.
    \item[Type reward $R^t_{\mathrm{type}}$.] It checks whether the type of the current action matches the ground truth action type. 
    If the types do not match, the reward is set to zero and subsequent rewards for this step are not computed. If the types match, a reward of 1 is assigned for this component.    
    \item[Necessary/Optional parameter reward] \textbf{$R^t_{\mathrm{nec}}$/ $R^t_{\mathrm{opt}}$.}  We categorize the parameters for all operations into two groups: \emph{necessary} and \emph{optional}. 
    Necessary parameters define the core functionality of an operation and allow minimal variation, whereas optional parameters are less critical, and differences may not affect the operation’s functionality.
    We evaluate the necessary and optional groups independently. For each group, we compute a score in the range [0, 1] as the average matching quality of the parameters within that group.
\end{description}

Overall, the accuracy reward for the $t$-th action $R^t_{\mathrm{acc}}$, is computed as:
\begin{equation*}
    R^t_{\mathrm{acc}} =  R^t_{\mathrm{format}} + R^t_{\mathrm{type}} + R^t_{\mathrm{necc}} + R^t_{\mathrm{opt}}.
\end{equation*}

In addition to this accuracy reward, we introduce two auxiliary rewards to further stabilize the RL training process and mitigate potential reward hacking during the RLVR phase.

The first auxiliary reward is designed to penalize actions that exceed the ground‑truth sequence. In our current formulation, such over‑predicted actions receive a fixed negative reward.
However, setting this penalty too small or too large can bias the model toward producing overly long or overly short action sequences.
To address this, we propose an adaptive penalty scheme for these exceeding actions. 
Specifically, we aggregate the rewards from all non-exceeding actions at the same time step across the other samples in the batch. We then assign penalties to the exceeding actions such that the average reward for predictions at the current step is zero across the entire batch. Formally, the exceeding reward for sample $i$ at step $t$ is calculated as:
\begin{equation*}
    R^t_{i, \mathrm{exc}} =
    \begin{cases}
    - \frac{1}{N^t_{\mathrm{exc}}}\sum_{j\in\mathcal{B}^t} R^t_{j,\mathrm{acc}}, & i \notin \mathcal{B}^t, \\
    0, &i \in \mathcal{B}^t,
    \end{cases}
\end{equation*}
where $N^t_{exc}$ is the number of samples in the current batch where the model predicts an exceeding action at step $t$, while $\mathcal{B}^t$ denotes the set of the samples in the current batch whose ground truth action sequence has at least $t$ steps.

The second auxiliary reward is the action type distribution modifier. 
Due to the varying difficulty in achieving rewards across different action types, the model may develop an undesirable bias towards predicting types that are easier to reward. To address this issue, we introduce a reward modifier based on the distributional differences of the action types.
Specifically, we calculate the distribution of action types in the ground-truth annotations and in the model predicted action sequences within the current batch, respectively. The distribution modifier for the action at step $t$ is then calculated as:
\begin{equation*}
    R^t_{\mathrm{dist}} =
    \begin{cases}
        \frac{p^{t}_{\mathrm{gt}}\, - \, p^{t}_{\mathrm{pred}}}{\max{(p^{t}_{\mathrm{gt}},\, p^{t}_{\mathrm{pred}})}},&  \frac{p^{t}_{\mathrm{gt}}\, - \,p^{t}_{\mathrm{pred}}}{\max{(p^{t}_{\mathrm{gt}},\, p^{t}_{\mathrm{pred}})}} > \theta, \\
        0,&  \text{otherwise},
    \end{cases}
\end{equation*}
where $p^{t}_{\mathrm{gt}}$ denotes the proportion of ground‑truth actions in the current batch that share the same type as the predicted action at step $t$.
Similarly, $p^{t}_{\mathrm{pred}}$ denotes the proportion of predicted actions in the current batch that belong to this action type, and $\theta$ is a predefined threshold.

Finally, the total reward for the action at step $t$ used during training is given by:
\begin{equation*}
    R^t = R^t_{\mathrm{acc}} + R^t_{\mathrm{exc}} + R^t_{\mathrm{dist}}.
\end{equation*}
A detailed algorithm describing the computation process of the final reward is presented in the Appendix.
Future versions will consider incorporating additional factors, such as enforcing consistency between the mass and the number of moles used.

\paragraph{PPO Training.}
Building on the designed reward function, we fine-tune the \reasoningmodel{} model using the Proximal Policy Optimization (PPO) algorithm\cite{schulman2017proximal}.
PPO is an empirically stable choice for language model alignment and complex reasoning tasks \cite{ouyang2022training, zeng2024token, wang2025reinforcement}, particularly well-suited for our dense reward setting.
The PPO objective function is defined as:
\begin{equation*}  
\mathcal{L}_{\text{PPO}}(\theta, \psi) = \mathcal{L}_{\text{clip}}(\theta) - w_1 \mathrm{KL}(\theta) - w_2 \mathcal{L}(\psi),  
\end{equation*}
where $\theta$ represents the language model parameters, and $\psi$ represents the critic model parameters. 
The clipped objective, $\mathcal{L}_{\text{clip}}(\theta)$, maximizes the expected reward while restricting large policy updates. 
The KL penalty $\mathrm{KL}(\theta)$ limits divergence from the reference policy (e.g., the SFT model). 
Lastly, the critic loss, $\mathcal{L}(\psi)$, minimizes value estimation errors, improving reward prediction and overall training stability.

\begin{table*}[ht]  
\centering  
\caption{Performance comparison of \reasoningmodel~against baselines on traditional NLP metrics. Metrics are in the interval [0, 100].}  
\label{tab:nlp_metrics}  
  
\begin{adjustbox}{max width=\textwidth}  
\begin{tabular}{l  
                S[table-format=2.1] S[table-format=2.1] 
                S[table-format=2.1] S[table-format=1.1] S[table-format=1.1] S[table-format=2.1] 
                S[table-format=2.1] S[table-format=2.1] S[table-format=2.1] 
                S[table-format=2.1] S[table-format=2.1]} 
\toprule  
\textbf{Model} & \textbf{BLEU-2} & \textbf{BLEU-4} & \textbf{LEV avg} & \textbf{LEV 90\%} & \textbf{LEV 75\%} & \textbf{LEV 50\%} & \textbf{Rouge-1} & \textbf{Rouge-2} & \textbf{Rouge-L} & \textbf{METEOR} & \textbf{Seq-O} \\  
\midrule  
Nearest Neighbor        & 54.5 & 39.1 & 46.6 & \textbf{1.0} & 2.4 & 30.2 & 60.0 & 38.4 & 47.3 & 50.0 & 65.1 \\  
1-shot GPT-4o           & 39.6 & 29.9 & 31.3 & 0.0 & 0.0 &  1.4 & 50.4 & 26.2 & 36.5 & 55.1 &  6.1 \\  
3-shot GPT-4o           & 41.5 & 31.8 & 32.7 & 0.0 & 0.0 &  2.4 & 53.1 & 30.1 & 39.4 & 57.4 &  8.4 \\  
1-shot o4-mini (high)   & 55.2 & 43.8 & 42.7 & 0.0 & 0.1 & 17.6 & 60.4 & 36.9 & 46.6 & 58.9 & 26.6 \\  
3-shot o4-mini (high)   & 56.8 & 45.5 & 44.7 & 0.0 & 0.3 & 24.0 & 62.9 & 39.9 & 49.2 & 61.5 & 26.2 \\  
1-shot GPT-5 (high)     & 67.1 & 56.4 & 51.5 & 0.8 & 2.9 & 54.1 & 67.6 & 49.4 & 55.3 & 64.7 & 59.9 \\  
3-shot GPT-5 (high)     & 65.0 & 54.4 & 51.7 & 0.6 & 2.6 & 55.4 & 68.5 & 50.1 & 55.9 & 66.9 & 61.1 \\  
\midrule  
\reasoningmodel{}-8B (SFT) & 69.9 & 59.4 & 56.8 & 0.1 & 2.8 & 77.9 & 71.2 & 52.9 & 59.8 & 68.6 & 69.4 \\  
\reasoningmodel{}-32B (SFT) & 70.2 & 59.7 & 57.2 & 0.1 & 3.2 & 78.8 & 71.5 & 53.3 & 60.2 & 68.8 & 70.1 \\  
\reasoningmodel{}-8B (RL)   & \textbf{72.0} & \textbf{61.3} & \textbf{57.4} & 0.3 & \textbf{4.3} & \textbf{78.8} & \textbf{72.1} & \textbf{54.5} & \textbf{61.1} & \textbf{69.7} & \textbf{70.9} \\  
\bottomrule  
\end{tabular}  
\end{adjustbox}  
\end{table*}

\begin{figure*}[ht]
    \centering
    \includegraphics[width=0.9\textwidth]{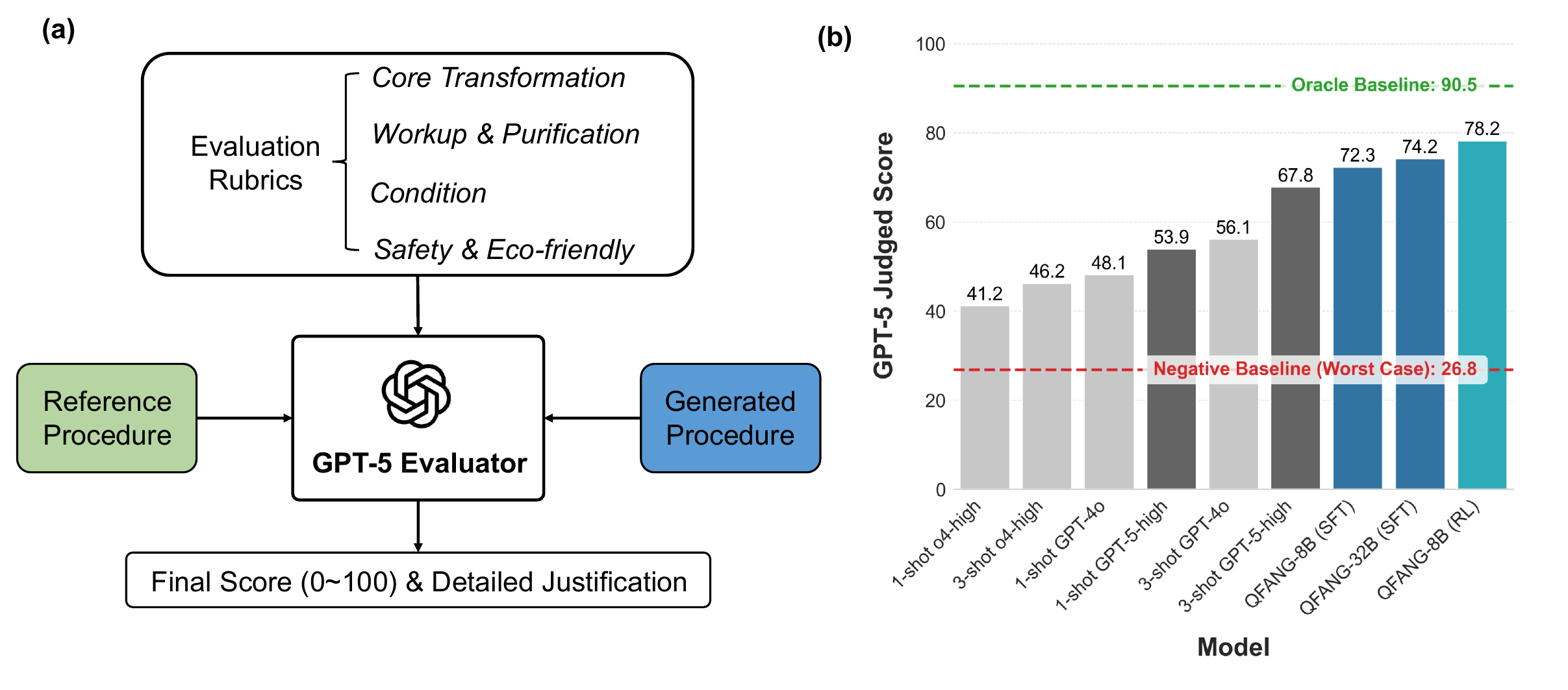}
    \caption{LLM-as-a-judge evaluator and performance comparison.
    (a) Overview of the evaluation process, where the GPT-5 judge receives the reference procedure, the generated procedure, and a detailed scoring rubric. (b) Bar chart comparing \reasoningmodel{}’s final scores with baseline models; horizontal dashed lines mark the Oracle baseline (upper) and the lowest-performing negative baseline (lower).
    }
    \label{fig:llm_judge_results}
\end{figure*}

\paragraph{GRPO Training.} 
Alternatively, the dense reward structure can be reformulated as  
an outcome‑based setting, allowing the use of Group Relative Policy Optimization (GRPO) \cite{shao2024deepseekmath} to fine-tune \reasoningmodel{}.
The GRPO objective function is defined as:
\begin{equation*}  
\mathcal{L}_{\text{GRPO}}(\theta) = \mathcal{L}_{\text{clip}}(\theta) - w_1 \mathrm{KL}(\theta). 
\end{equation*}
A key innovation of GRPO over PPO is the elimination of a separate critic model. Instead, GRPO estimates the advantage by comparing the reward of each response to the average reward obtained from a group of responses generated by the same policy.
Given a question $x$ from the dataset, GRPO samples $G$ responses $y_1, \dots, y_G$, and assigns rewards $r_1, \dots, r_G$ to each. For the $i$-th response in the group, the corresponding group‑relative advantage is computed as:
\begin{equation*}
\hat{A}_{i} = \frac{r_i - \text{mean}(r_1, \dots, r_G)}{\text{std}(r_1, \dots, r_G)}.
\end{equation*}

Additional details on the loss components are provided in the Appendix.

\section*{Results}
\paragraph{Quantitative Evaluation: Benchmarking against Baselines.}
To assess the performance of \reasoningmodel{}, we conducted a comprehensive quantitative evaluation against several strong baselines using standard text-similarity metrics, along with a new LLM-as-a-judge assessment.

To contextualize the performance of \reasoningmodel{}, we established a set of competitive baselines, comprising both retrieval-based and advanced generative approaches. The simplest baseline is a non-generative Nearest Neighbor (NN) method. For a given target reaction, this approach identifies the single most similar reaction from the training set, based on Tanimoto similarity of their reaction fingerprints (DRFP\cite{probst2022drfp}) and directly outputs its associated procedure as the prediction. We further tested against a suite of state-of-the-art LLMs, including GPT-4o, o4-mini (high) and GPT-5 (high), leveraging a retrieval-augmented in-context learning strategy. For each target reaction, we retrieved the top-$k$ most similar reactions and their ground-truth procedures from the training set using the same DRFP-based similarity metric. These retrieved pairs were then integrated into the prompt as in-context examples to guide the model's generation. We evaluated this strategy in both 1-shot ($k$=1) and 3-shot ($k$=3) settings. 
During inference, \reasoningmodel{} operates without in-context learning.

\subparagraph{Superior Performance on Traditional NLP Metrics.}
Similar to previous studies, we evaluated the lexical similarity between the generated and reference procedures using a suite of established metrics\cite{smiles2actions, chen2025reactgpt}, with results presented in Table~\ref{tab:nlp_metrics}. These included BLEU, a metric based on n-gram precision originally developed for machine translation\cite{papineni2002bleu}; ROUGE, which measures n-gram recall\cite{lin2004rouge}; and METEOR, which incorporates synonyms and stemming for more robust alignment\cite{banerjee2005meteor}. We also report metrics based on normalized Levenshtein (LEV) similarity, which quantifies character-level edit distance\cite{levenshtein}; specifically, LEV X\% denotes the fraction of predictions achieving a normalized similarity score of X\% or greater when compared to the reference\cite{smiles2actions}.
Finally, we include Seq-O, a metric designed to measure the similarity of the core action verb sequences, providing insight into the model's ability to capture the procedural workflow\cite{chemtrans}.

Across all evaluation metrics, \reasoningmodel{} consistently outperforms all baseline models.
The RL trained variant, \reasoningmodel{}-8B (RL), achieves a BLEU-4 score of 61.3 and a ROUGE-L score of 61.1. 
This substantially surpasses the strongest generative baseline, 3-shot GPT-5 (high), which scored 54.4 and 55.9, respectively. This trend holds for our SFT models as well; the \reasoningmodel{}-32B (SFT) variant reached a METEOR score of 68.8 and a Seq-O score of 70.1, indicating a strong capability to generate text that is both syntactically correct and lexically aligned with the ground-truth data. Notably, the performance of retrieval-augmented LLMs is highly dependent on the number of in-context examples, increasing the number of examples does not necessarily lead to better scores.
We also find that the simple Nearest Neighbor baseline performs well on some metrics like BLEU-2 and Seq-O, suggesting that many reactions in the test set may have close analogues in the training data.

To further investigate whether the superior average performance of \reasoningmodel{} stems from true generalization or merely an improved ability to recall similar training examples, we conducted a more rigorous analysis. We stratified the test set based on its procedural and chemical similarity to the training data. The results, detailed in Appendix, demonstrate that \reasoningmodel{} maintains its performance advantage even on samples that are highly dissimilar to the training set, whereas the baseline models exhibit a much sharper decline. This provides strong evidence that \reasoningmodel{} has learned to reason from underlying chemical principles, validating the effectiveness of our CGR training methodology.

Despite the superior performance of \reasoningmodel{} on these metrics, we posit that such surface-level comparisons are insufficient for rigorously evaluating chemical procedure generation. Lexical metrics are inherently insensitive to chemical logic and safety. For example, a simple string comparison would penalize the substitution of a hazardous reagent like sodium hydride (\ce{NaH}) with a milder base like sodium hydroxide (\ce{NaOH}) with only a minor score reduction, yet this represents a fundamental and potentially dangerous misunderstanding of the required reaction conditions. Similarly, a chemically consistent substitution, such as writing \ce{NEt3} instead of triethylamine, would be incorrectly penalized as a mismatch. While the strong performance of \reasoningmodel{} on these metrics is a prerequisite, a more meaningful assessment of chemical soundness and procedural viability is essential for a comprehensive evaluation of \reasoningmodel{}’s capabilities. This motivates our development of a chemically-aware, LLM-based evaluation framework.

\paragraph{LLM-as-a-Judge: A Chemically-Aware Evaluator.}

\begin{figure*}[h]
    \centering
    \includegraphics[width=\textwidth]{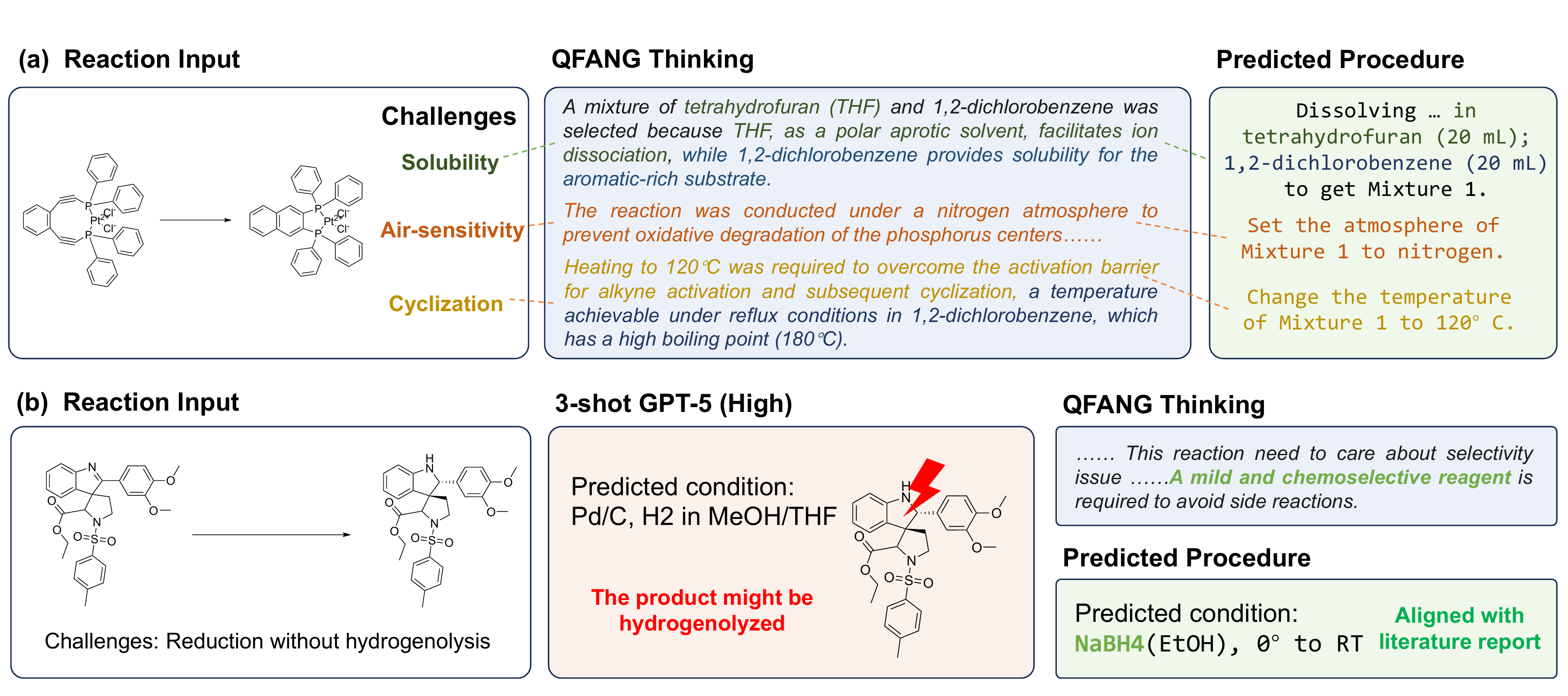}
    \caption{
    Demonstration of \reasoningmodel{}'s ability to generalize across diverse chemical contexts.
    (a) For an out-of-distribution organometallic reaction, \reasoningmodel{} infers a viable synthetic procedure from first principles. It correctly identifies structural challenges like poor solubility and air-sensitivity, and proposes expert-level solutions (e.g., a binary solvent system and an inert atmosphere) in its generated procedure. (b) Achieving chemoselectivity on natural product intermediates. When tasked with a sensitive imine reduction,
    \reasoningmodel{} exhibits superior chemical judgment. In contrast to a 3-shot GPT-5 baseline, which suggests harsh and potentially destructive conditions (\ce{Pd/C}, \ce{H2}), \reasoningmodel{} selects a mild and chemoselective reagent (\ce{NaBH4}), thereby preventing side reactions and aligning with the literature-reported gold standard.}
    \label{fig:generalizability}
\end{figure*}

To address the semantic gap of traditional NLP metrics, we designed and implemented a more rigorous evaluation framework that leverages an expert-level LLM, GPT-5 (high), as a judge. This framework moves beyond simple lexical comparison to assess generated procedures on their chemical viability, procedural completeness, and strict adherence to a machine-readable format, reflecting the practical requirements for synthesis planning and eventual laboratory automation.

The evaluation of each generated procedure is performed by providing the GPT‑5 judge with a composite prompt that includes the ground‑truth reference procedure (as the gold standard), the model‑generated procedure, and a detailed scoring rubric. The rubric consists of four categories: \textit{Reaction Score} (40 points), assessing the core transformation and stoichiometry; \textit{Workup and Purification Score} (30 points), evaluating the separation process; \textit{Conditions Score (20 points)}, examining solvents and reagents; and \textit{Safety and Modern Practice Score} (10 points), penalizing usage of outdated and toxic components. The prompt also penalized deviations from a required machine-readable syntax.

To validate our LLM-as-a-judge framework and contextualize the scores, we established several control baselines designed to probe the judge's sensitivity to specific types of procedural errors. We created three \textit{negative baselines} by systematically corrupting the ground-truth procedures: (1) \textit{Reagent}, where a key reagent was replaced with a chemically nonsensical alternative; (2) \textit{Swap Actions}, where the order of two critical steps was inverted; and (3) \textit{Both}, which combined both errors. Conversely, we established an \textit{Oracle baseline} by making only chemically benign modifications to the ground-truth procedures, such as replacing reagents with known synonyms (e.g., substituting methanol with MeOH). As shown in Figure~\ref{fig:llm_judge_results}, these baselines provide a clear scale for interpretation. 
The \textit{Oracle baseline} achieves a near-perfect score of 90.5, confirming the judge's ability to tolerate reasonable chemical variations. In contrast, the \textit{negative baselines} score significantly lower, dropping to 39.1, 39.7, and 26.8, respectively, demonstrating the framework's capability to effectively penalize chemically and structurally flawed procedures.

The results of this expert-level LLM evaluation, presented in Figure~\ref{fig:llm_judge_results}, further solidify the superior capabilities of \reasoningmodel{}.
Our best-performing model, \reasoningmodel{}-8B (RL), achieves a final score of 78.2, surpassing the strongest baseline (3-shot GPT-5-high at 67.8) by a substantial margin and closely approaching the Oracle score. This high score indicates that the procedures generated by \reasoningmodel{} are not only chemically sound and logical but also adhere precisely to the stringent formatting required for robust, automated laboratory systems. The performance of the SFT models, \reasoningmodel{}-8B (SFT) and \reasoningmodel{}-32B (SFT) at 72.3 and 74.2, respectively, also shows a clear advantage over the retrieval-augmented LLMs. This outcome validates our training methodology, demonstrating that our model has learned the underlying principles of chemical procedure design rather than merely mimicking surface-level text patterns.

\begin{figure*}[h]
    \centering
    \includegraphics[width=1.0\textwidth]{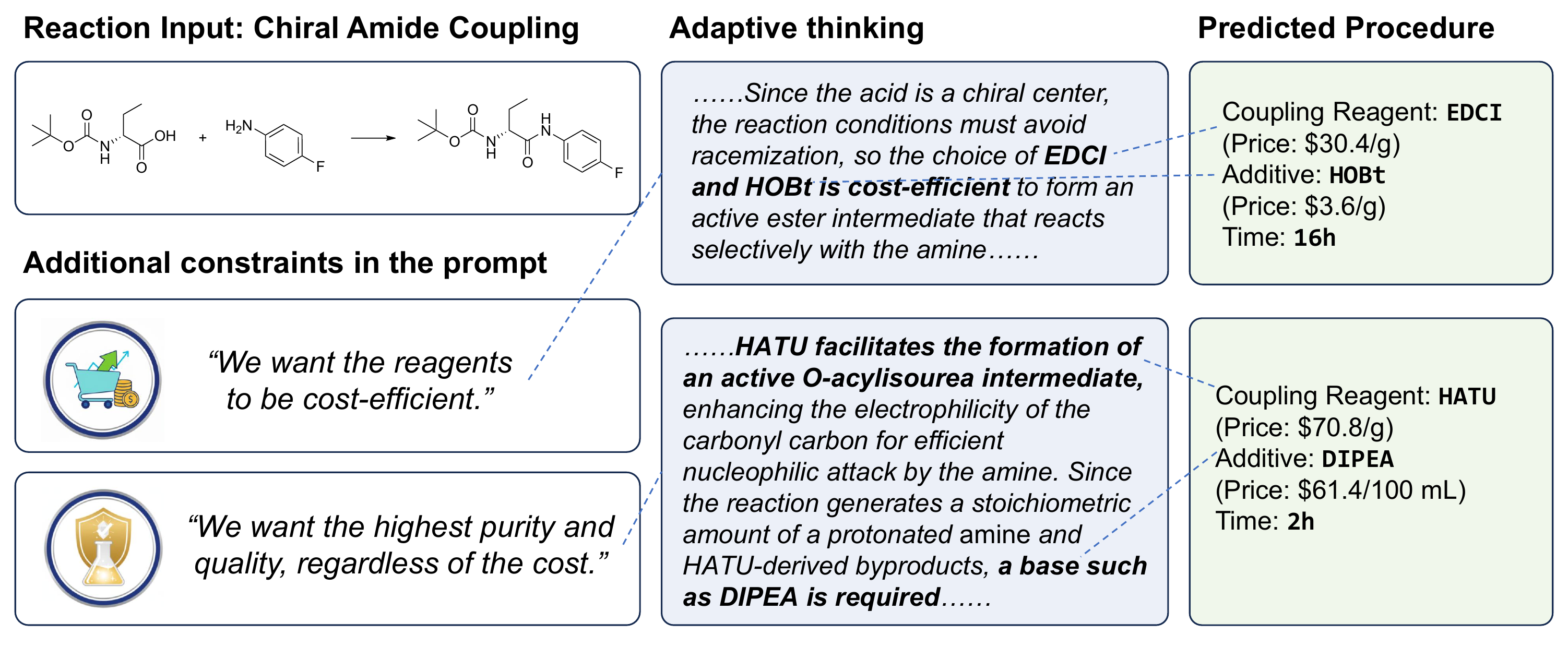}
    \caption{
    Demonstration of \reasoningmodel{}'s ability to adapt based on user-specific constraints. For the same chiral amide coupling reaction, the model proposes two distinct, chemically valid procedures, each tailored to different and potentially competing user objectives: (i) cost efficiency, and (ii) maximum purity and quality.}
    \label{fig:adaptivity_case}
\end{figure*}

\paragraph{Case Studies}
\subparagraph{Generalizability: Navigating Out-of-Domain Chemical Challenges.}

\begin{figure*}[h]
    \centering \includegraphics[width=0.9\textwidth]{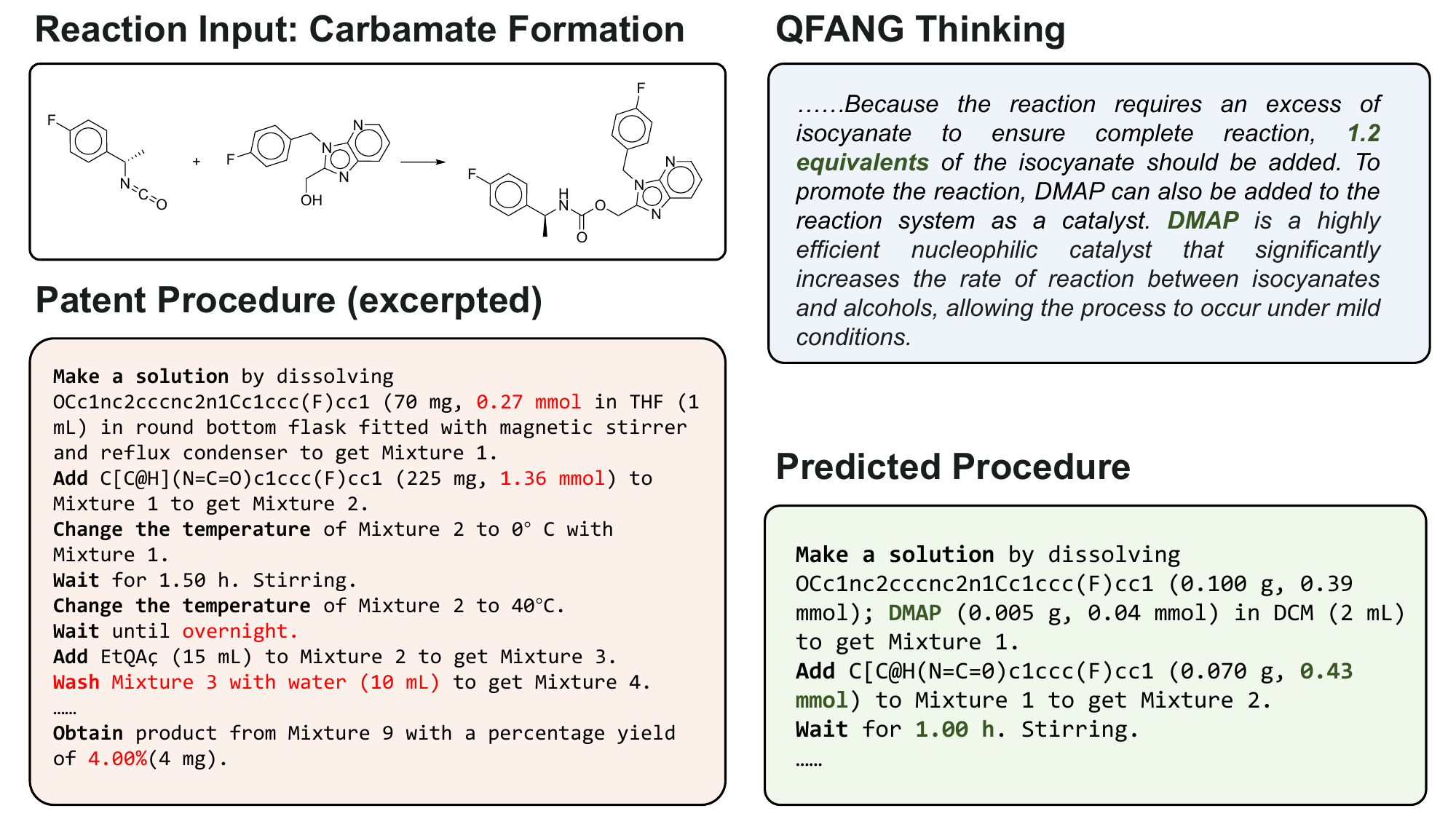}
    \caption{
    Demonstration of  \reasoningmodel{}’s ability to correct flawed procedures. When presented with a stoichiometrically imbalanced reaction from the source dataset, the model produces a chemically valid and optimized alternative, incorporating the appropriate catalyst to ensure reaction feasibility.}\label{fig:rectification_case}
\end{figure*}
We evaluated the model's generalization ability through two distinct and highly challenging cases: one requiring reasoning in a completely novel chemical domain, and another demanding nuanced selectivity within the familiar context of natural product synthesis.

Our first challenge tested the model's ability to reason from first principles on a complex, out-of-distribution reaction from the domain of organophosphorus and organometallic chemistry, a field far removed from the drug synthesis patents that constitute its training data. The task was to devise a procedure for an intramolecular cycloaromatization within a platinum-diphosphine heterocyclic framework~(Figure~\ref{fig:generalizability}(a))\cite{lindahl2025pronounced}. Success in this task hinges not on recalling a known reaction template, but on a fundamental analysis of the substrate's unique structural features: its poor solubility, the presence of air-sensitive phosphorus centers, and the high activation energy typical of such bond reorganization reactions.

Remarkably, \reasoningmodel{} generated a highly plausible and expert-level procedure. It correctly inferred the need for a binary solvent system, pairing a polar aprotic solvent (\texttt{THF}) to solvate the ionic phosphonium portions with a high-boiling-point, nonpolar aromatic solvent (\texttt{1,2-dichlorobenzene}) to dissolve the hydrophobic backbone and provide the necessary thermal energy. Furthermore, it identified the air-sensitive nature of the phosphorus centers and correctly prescribed an inert nitrogen atmosphere. The proposed conditions of moderate-high temperature (\texttt{120°C}) and prolonged reaction time (\texttt{overnight}) are also fully consistent with the high activation barrier expected for such a Bergman cycloaromatization. This case demonstrates a profound capability to deconstruct a novel molecular structure and devise a viable experimental plan based on core chemical principles of solubility, stability, and reactivity.

Having established the model's ability to navigate a new chemical domain, we next probed its capacity to handle fine-grained selectivity within a more familiar, yet notoriously complex area: natural product synthesis. The second challenge was a critical reduction step from R.B. Woodward's landmark total synthesis of strychnine \cite{woodward1954strychnine}. The task involves the reduction of an imine within the dense, polyfunctionalized core of the natural product~(Figure~\ref{fig:generalizability}(b)). The primary difficulty lies in achieving chemoselectivity, as a non-selective or overly harsh reducing agent could easily lead to undesired side reactions, such as the hydrogenolysis of the newly formed C-N bond.

As shown in Figure~\ref{fig:generalizability}, \reasoningmodel{} again successfully navigated this challenge by proposing the use of sodium borohydride, a mild and selective reducing agent. This prediction aligns perfectly with the established literature procedure, achieving the desired clean conversion. In stark contrast, a powerful baseline like GPT-5-high, when prompted, proposed catalytic hydrogenation with Palladium on carbon. While effective for simple imines, these powerful conditions are ill-suited for this delicate substrate and would likely cause product decomposition.
This pair of case studies effectively illustrates the breadth and depth of \reasoningmodel{}’s chemical understanding, showcasing its ability to reason from context-specific rules in complex domains, which is essential for reliable real-world synthesis planning.
Beyond chemical novelty, we also evaluated operational generalization. In the Appendix, we demonstrate \reasoningmodel{}'s ability to transition from discovery-scale to process-scale chemistry, successfully designing a chromatography-free, 50-kg manufacturing procedure under industrial constraints.

\subparagraph{Goal-Oriented Planning with User Constraints.}

A key feature of a practical synthesis planning system is its ability to adapt and re-optimize plans based on user-specified high-level constraints. 
We illustrate this through a case study of chiral amide coupling, a common transformation where maintaining stereochemical integrity is crucial. This example highlights that there is no single "best" procedure; rather, the optimal conditions are determined by a nuanced trade-off between cost, time, and the desired stereochemical purity\cite{albeiicio2001peptide}.

As illustrated in Figure~\ref{fig:adaptivity_case}, we prompted \reasoningmodel{} with the same chemical transformation but under two distinct, competing constraints. Initially, when prompted with the constraint \textit{“We want the reagents to be cost-efficient.”}, \reasoningmodel{} generated a procedure centered on the use of EDCI and HOBt. The reasoning trace identified the substrate's chirality and the inherent risk of racemization associated with simple carbodiimide activators. Consequently, it correctly included HOBt as a necessary additive to suppress this side reaction, proposing a 16 hour procedure that reflects a standard, cost-effective method suitable for a basic laboratory setup.

In a subsequent interaction, the model was given a new, quality-focused directive: \textit{“We want the highest purity and quality, regardless of the cost.”}. In response, \reasoningmodel{} dynamically altered its plan, replacing the previous system with the more expensive but superior coupling reagent, HATU, and the base DIPEA. The reasoning trace shifted to prioritize efficiency and stereochemical fidelity, explaining that modern uronium salt-based reagents like HATU are renowned for their rapid kinetics and their exceptional ability to prevent racemization. This was reflected in the final procedure, which reduced the reaction time from 16 hours to just 2 hours while ensuring the highest possible enantiomeric purity—a standard expected in an advanced R\&D setting where speed and quality are paramount.

This case study illustrates that the understanding of \reasoningmodel{} goes beyond surface-level reagent selection. It grasps the deep, practical trade-offs in synthetic chemistry and can select the appropriate methodology based on high-level user goals, demonstrating a crucial step towards creating a collaborative and intelligent tool for synthesis planning.

\subparagraph{Surpassing and Correcting Flawed Training Data.}
An interesting finding from our study is that \reasoningmodel{} can identify and correct chemically flawed procedures present in its training data. This capability indicates that the model has internalized fundamental chemical principles, allowing it to generate procedures that are superior to some of the examples it was trained on.  Its function as a corrective filter is critical for building reliable systems from vast but imperfect data sources, such as chemical patents. We illustrate this with a representative case from the training set, where the original patented procedure is stoichiometrically questionable, leading to a very low reported yield.

The case involves an acylation reaction where the original procedure calls for a four-fold excess of a highly reactive isocyanate~(Figure~\ref{fig:rectification_case}). From a chemical standpoint, such a large excess is not only wasteful but also detrimental to the process. The unreacted electrophile will inevitably undergo hydrolysis during the aqueous workup, leading to significant byproduct formation (e.g., urea) that complicates purification and ultimately contributes to the low reported yield of just 4\%\cite{stockley2020autotaxinpatent}.

Instead of merely replicating this flawed method, \reasoningmodel{} proposes a refined and more efficient procedure. Its CoT analysis correctly identifies the core transformation but implicitly rejects the erroneous stoichiometry. The model generates a new protocol that reduces the isocyanate to a more standard 1.2 equivalents, a quantity sufficient to drive the reaction to completion without creating an excessive purification burden. Furthermore, it introduces a catalytic amount of DMAP (4-Dimethylaminopyridine) to ensure an efficient reaction, a common best practice for acylations that was absent in the original text. Finally, it implements a more robust, multi-step basic and neutral wash protocol designed to effectively remove both acidic and neutral impurities before the final purification.

This case study shows that \reasoningmodel{} does not blindly reproduce the flawed methods present in its training corpus. Instead, it leverages its learned understanding of stoichiometry and reaction optimization to generate a chemically sound and superior alternative. This ability to critically assess and rectify suboptimal procedures underscores its potential as a tool not only for novel synthesis planning but also for the crucial validation and refinement of existing experimental protocols.

\section*{Conclusion}
In this work, we addressed the critical gap between synthesis planning and machine-readable laboratory actions by introducing \reasoningmodel{}, a scientific reasoning model capable of generating high-fidelity chemical procedures. 
Our approach was built on three key pillars: 
(i) construction of a large-scale, structured procedure dataset through a novel LLM-based automatic annotation pipeline; 
(ii) development of a chemistry-guided reasoning framework to elicit chemical-principle-based reasoning in the model; 
and (iii) integration of a reinforcement learning with verifiable rewards stage to further enhance predictive accuracy.

Comprehensive evaluations demonstrate that \reasoningmodel{} achieves competitive performance on established NLP related metrics, surpassing strong baselines such as retrieval-augmented GPT-5, while maintaining high chemical validity under the LLM-as-a-judge assessment framework.
Qualitative analyses further reveal the emergence of chemical reasoning capabilities, as reflected in the model’s ability to generalize to out-of-domain reactions, adapt experimental plans to high-level user constraints, and even identify and amend flawed procedures within its own training data. 
By generating robust, well-reasoned, and machine-readable experimental protocols, \reasoningmodel{} represents a promising step toward bridging the loop between in-silico design and laboratory execution, contributing to the progress of next-generation autonomous platforms for scientific discovery.

\section*{Acknowledgements}
We gratefully acknowledge the Microsoft Research AI for Science team for their invaluable discussions, feedback, and support. In particular, we would like to thank (in alphabetical order): Rianne van den Berg, Yeqi Bai, Andrew Fowler, Deniz Gunceler, Jean H\'elie, Jos\'e Jim\'enez-Luna, Bonnie Kruft, Elise van der Pol, Robert Pinsler, Maik Riechert, Yu Xie, Tian Xie, and Claudio Zeni for proofreading and valuable feedback.
We also thank Xinnuo Xu, Ryota Tomioka and Katja Hofmann from Microsoft Research Cambridge for their helpful discussions and feedback.

\bibliographystyle{naturemag}
\bibliography{main}

@article{vaucher2020automated,
  title={Automated extraction of chemical synthesis actions from experimental procedures},
  author={Vaucher, Alain C and Zipoli, Federico and Geluykens, Joppe and Nair, Vishnu H and Schwaller, Philippe and Laino, Teodoro},
  journal={Nature communications},
  volume={11},
  number={1},
  pages={3601},
  year={2020},
  publisher={Nature Publishing Group UK London}
}

@inproceedings{llm_hallucinate,
  title={Llms will always hallucinate, and we need to live with this},
  author={Banerjee, Sourav and Agarwal, Ayushi and Singla, Saloni},
  booktitle={Intelligent Systems Conference},
  pages={624--648},
  year={2025},
  organization={Springer}
}

@article{qwen3,
  title={Qwen3 technical report},
  author={Yang, An and Li, Anfeng and Yang, Baosong and Zhang, Beichen and Hui, Binyuan and Zheng, Bo and Yu, Bowen and Gao, Chang and Huang, Chengen and Lv, Chenxu and others},
  journal={arXiv preprint arXiv:2505.09388},
  year={2025}
}

@article{gpt4,
  title={Gpt-4 technical report},
  author={Achiam, Josh and Adler, Steven and Agarwal, Sandhini and Ahmad, Lama and Akkaya, Ilge and Aleman, Florencia Leoni and Almeida, Diogo and Altenschmidt, Janko and Altman, Sam and Anadkat, Shyamal and others},
  journal={arXiv preprint arXiv:2303.08774},
  year={2023}
}

@article{localmapper,
  title={Precise atom-to-atom mapping for organic reactions via human-in-the-loop machine learning},
  author={Chen, Shuan and An, Sunggi and Babazade, Ramil and Jung, Yousung},
  journal={Nature Communications},
  volume={15},
  number={1},
  pages={2250},
  year={2024},
  publisher={Nature Publishing Group UK London}
}

@phdthesis{lowe2012extraction,
  title={Extraction of chemical structures and reactions from the literature},
  author={Lowe, Daniel Mark},
  school={University of Cambridge},
  year={2012}
}

@inproceedings{zhong2024actionie,
  title={Actionie: Action extraction from scientific literature with programming languages},
  author={Zhong, Xianrui and Du, Yufeng and Ouyang, Siru and Zhong, Ming and Luo, Tingfeng and Ho, Qirong and Peng, Hao and Ji, Heng and Han, Jiawei},
  booktitle={Proceedings of the 62nd Annual Meeting of the Association for Computational Linguistics (Volume 1: Long Papers)},
  pages={12656--12671},
  year={2024}
}

@article{liu2024reactxt,
  title={Reactxt: Understanding molecular" reaction-ship" via reaction-contextualized molecule-text pretraining},
  author={Liu, Zhiyuan and Shi, Yaorui and Zhang, An and Li, Sihang and Zhang, Enzhi and Wang, Xiang and Kawaguchi, Kenji and Chua, Tat-Seng},
  journal={arXiv preprint arXiv:2405.14225},
  year={2024}
}

@article{smiles2actions,
  title={Inferring experimental procedures from text-based representations of chemical reactions},
  author={Vaucher, Alain C and Schwaller, Philippe and Geluykens, Joppe and Nair, Vishnu H and Iuliano, Anna and Laino, Teodoro},
  journal={Nature communications},
  volume={12},
  number={1},
  pages={2573},
  year={2021},
  publisher={Nature Publishing Group UK London}
}

@article{chemtrans,
  title={Transcription between human-readable synthetic descriptions and machine-executable instructions: an application of the latest pre-training technology},
  author={Zeng, Zheni and Nie, Yi-Chen and Ding, Ning and Ding, Qian-Jun and Ye, Wei-Ting and Yang, Cheng and Sun, Maosong and Zhu, Rong and Liu, Zhiyuan and others},
  journal={Chemical Science},
  volume={14},
  number={35},
  pages={9360--9373},
  year={2023},
  publisher={Royal Society of Chemistry}
}

@inproceedings{papineni2002bleu,
  title={Bleu: a method for automatic evaluation of machine translation},
  author={Papineni, Kishore and Roukos, Salim and Ward, Todd and Zhu, Wei-Jing},
  booktitle={Proceedings of the 40th annual meeting of the Association for Computational Linguistics},
  pages={311--318},
  year={2002}
}

@inproceedings{christofidellis2023unifying,
  title={Unifying molecular and textual representations via multi-task language modelling},
  author={Christofidellis, Dimitrios and Giannone, Giorgio and Born, Jannis and Winther, Ole and Laino, Teodoro and Manica, Matteo},
  booktitle={International Conference on Machine Learning},
  pages={6140--6157},
  year={2023},
  organization={PMLR}
}

@inproceedings{lin2004rouge,
  title={Rouge: A package for automatic evaluation of summaries},
  author={Lin, Chin-Yew},
  booktitle={Text summarization branches out},
  pages={74--81},
  year={2004}
}

@inproceedings{banerjee2005meteor,
  title={METEOR: An automatic metric for MT evaluation with improved correlation with human judgments},
  author={Banerjee, Satanjeev and Lavie, Alon},
  booktitle={Proceedings of the acl workshop on intrinsic and extrinsic evaluation measures for machine translation and/or summarization},
  pages={65--72},
  year={2005}
}

@article{levenshtein,
  title={A normalized Levenshtein distance metric},
  author={Yujian, Li and Bo, Liu},
  journal={IEEE transactions on pattern analysis and machine intelligence},
  volume={29},
  number={6},
  pages={1091--1095},
  year={2007},
  publisher={IEEE}
}

@inproceedings{chen2025reactgpt,
  title={ReactGPT: Understanding of Chemical Reactions via In-Context Tuning},
  author={Chen, Zhe and Fang, Zhe and Tian, Wenhao and Long, Zhaoguang and Sun, Changzhi and Chen, Yuefeng and Yuan, Hao and Li, Honglin and Lan, Man},
  booktitle={Proceedings of the AAAI Conference on Artificial Intelligence},
  volume={39},
  pages={84--92},
  year={2025}
}

@article{probst2022drfp,
  title={Reaction classification and yield prediction using the differential reaction fingerprint DRFP},
  author={Probst, Daniel and Schwaller, Philippe and Reymond, Jean-Louis},
  journal={Digital discovery},
  volume={1},
  number={2},
  pages={91--97},
  year={2022},
  publisher={Royal Society of Chemistry}
}

@article{blakemore2018organicsynthesis,
  title={Organic synthesis provides opportunities to transform drug discovery},
  author={Blakemore, David C and Castro, Luis and Churcher, Ian and Rees, David C and Thomas, Andrew W and Wilson, David M and Wood, Anthony},
  journal={Nature chemistry},
  volume={10},
  number={4},
  pages={383--394},
  year={2018},
  publisher={Nature Publishing Group UK London}
}

@article{li2015synthesis,
  title={Synthesis of many different types of organic small molecules using one automated process},
  author={Li, Junqi and Ballmer, Steven G and Gillis, Eric P and Fujii, Seiko and Schmidt, Michael J and Palazzolo, Andrea ME and Lehmann, Jonathan W and Morehouse, Greg F and Burke, Martin D},
  journal={Science},
  volume={347},
  number={6227},
  pages={1221--1226},
  year={2015},
  publisher={American Association for the Advancement of Science}
}

@article{stanley2023fakeuntilmake,
  title={Fake it until you make it? Generative de novo design and virtual screening of synthesizable molecules},
  author={Stanley, Megan and Segler, Marwin},
  journal={Current Opinion in Structural Biology},
  volume={82},
  pages={102658},
  year={2023},
  publisher={Elsevier}
}

@article{ley2015organicsynthesis,
  title={Organic synthesis: march of the machines},
  author={Ley, Steven V and Fitzpatrick, Daniel E and Ingham, Richard J and Myers, Rebecca M},
  journal={Angewandte Chemie International Edition},
  volume={54},
  number={11},
  pages={3449--3464},
  year={2015},
  publisher={Wiley Online Library}
}

@article{szymkuc2016casp,
  title={Computer-assisted synthetic planning: the end of the beginning},
  author={Szymku{\'c}, Sara and Gajewska, Ewa P and Klucznik, Tomasz and Molga, Karol and Dittwald, Piotr and Startek, Micha{\l} and Bajczyk, Micha{\l} and Grzybowski, Bartosz A},
  journal={Angewandte Chemie International Edition},
  volume={55},
  number={20},
  pages={5904--5937},
  year={2016},
  publisher={Wiley Online Library}
}

@article{segler2018planning,
  title={Planning chemical syntheses with deep neural networks and symbolic AI},
  author={Segler, Marwin HS and Preuss, Mike and Waller, Mark P},
  journal={Nature},
  volume={555},
  number={7698},
  pages={604--610},
  year={2018},
  publisher={Nature Publishing Group UK London}
}

@article{li2023retroranker,
  title={RetroRanker: leveraging reaction changes to improve retrosynthesis prediction through re-ranking},
  author={Li, Junren and Fang, Lei and Lou, Jian-Guang},
  journal={Journal of Cheminformatics},
  volume={15},
  number={1},
  pages={58},
  year={2023},
  publisher={Springer}
}

@article{li2024retrobleu,
  title={Retro-BLEU: quantifying chemical plausibility of retrosynthesis routes through reaction template sequence analysis},
  author={Li, Junren and Fang, Lei and Lou, Jian-Guang},
  journal={Digital Discovery},
  volume={3},
  number={3},
  pages={482--490},
  year={2024},
  publisher={Royal Society of Chemistry}
}

@article{maziarz2025syntheseus,
  title={Re-evaluating retrosynthesis algorithms with syntheseus},
  author={Maziarz, Krzysztof and Tripp, Austin and Liu, Guoqing and Stanley, Megan and Xie, Shufang and Gai{\'n}ski, Piotr and Seidl, Philipp and Segler, Marwin HS},
  journal={Faraday Discussions},
  volume={256},
  pages={568--586},
  year={2025},
  publisher={Royal Society of Chemistry}
}

@inproceedings{maziarz2025chimera,
  title={Chemist-aligned retrosynthesis by ensembling diverse inductive bias models},
  author={Maziarz, Krzysztof and Liu, Guoqing and Tripp, Austin and Li, Junren and Gai{\'n}ski, Piotr and Segler, Marwin},
  booktitle={NeurIPS 2025 AI for Science Workshop}
}

@inproceedings{liu2023pdvn,
  title={Retrosynthetic planning with dual value networks},
  author={Liu, Guoqing and Xue, Di and Xie, Shufang and Xia, Yingce and Tripp, Austin and Maziarz, Krzysztof and Segler, Marwin and Qin, Tao and Zhang, Zongzhang and Liu, Tie-Yan},
  booktitle={International conference on machine learning},
  pages={22266--22276},
  year={2023},
  organization={PMLR}
}

@article{zhong2022rsmiles,
  title={Root-aligned SMILES: a tight representation for chemical reaction prediction},
  author={Zhong, Zipeng and Song, Jie and Feng, Zunlei and Liu, Tiantao and Jia, Lingxiang and Yao, Shaolun and Wu, Min and Hou, Tingjun and Song, Mingli},
  journal={Chemical Science},
  volume={13},
  number={31},
  pages={9023--9034},
  year={2022},
  publisher={Royal Society of Chemistry}
}

@article{lin2020automatic,
  title={Automatic retrosynthetic route planning using template-free models},
  author={Lin, Kangjie and Xu, Youjun and Pei, Jianfeng and Lai, Luhua},
  journal={Chemical science},
  volume={11},
  number={12},
  pages={3355--3364},
  year={2020},
  publisher={Royal Society of Chemistry}
}

@article{fang2023substructure,
  title={Single-step retrosynthesis prediction by leveraging commonly preserved substructures},
  author={Fang, Lei and Li, Junren and Zhao, Ming and Tan, Li and Lou, Jian-Guang},
  journal={Nature Communications},
  volume={14},
  number={1},
  pages={2446},
  year={2023},
  publisher={Nature Publishing Group UK London}
}

@article{segler2017neuralsym,
  title={Neural-symbolic machine learning for retrosynthesis and reaction prediction},
  author={Segler, Marwin HS and Waller, Mark P},
  journal={Chemistry--A European Journal},
  volume={23},
  number={25},
  pages={5966--5971},
  year={2017},
  publisher={Wiley Online Library}
}

@inproceedings{chen2020retrostar,
  title={Retro*: learning retrosynthetic planning with neural guided A* search},
  author={Chen, Binghong and Li, Chengtao and Dai, Hanjun and Song, Le},
  booktitle={International conference on machine learning},
  pages={1608--1616},
  year={2020},
  organization={PMLR}
}

@article{gao2018condition,
  title={Using machine learning to predict suitable conditions for organic reactions},
  author={Gao, Hanyu and Struble, Thomas J and Coley, Connor W and Wang, Yuran and Green, William H and Jensen, Klavs F},
  journal={ACS central science},
  volume={4},
  number={11},
  pages={1465--1476},
  year={2018},
  publisher={ACS Publications}
}

@article{sun2025multicondition,
  title={Data-driven recommendation of agents, temperature, and equivalence ratios for organic synthesis},
  author={Sun, Xiaoqi and Liu, Jiannan and Mahjour, Babak and Jensen, Klavs F and Coley, Connor W},
  journal={Chemical Science},
  year={2025},
  publisher={Royal Society of Chemistry}
}

@article{wang2025reacon,
  title={Reacon: a template-and cluster-based framework for reaction condition prediction},
  author={Wang, Zihan and Lin, Kangjie and Pei, Jianfeng and Lai, Luhua},
  journal={Chemical Science},
  volume={16},
  number={2},
  pages={854--866},
  year={2025},
  publisher={Royal Society of Chemistry}
}

@article{chen2024enhancingcondition,
  title={Enhancing chemical synthesis: a two-stage deep neural network for predicting feasible reaction conditions},
  author={Chen, Lung-Yi and Li, Yi-Pei},
  journal={Journal of Cheminformatics},
  volume={16},
  number={1},
  pages={11},
  year={2024},
  publisher={Springer}
}

@article{zhao2025chemdfm,
  title={Developing ChemDFM as a large language foundation model for chemistry},
  author={Zhao, Zihan and Ma, Da and Chen, Lu and Sun, Liangtai and Li, Zihao and Xia, Yi and Chen, Bo and Xu, Hongshen and Zhu, Zichen and Zhu, Su and others},
  journal={Cell Reports Physical Science},
  volume={6},
  number={4},
  year={2025},
  publisher={Elsevier}
}

@article{zhao2025chemdfmR,
  title={ChemDFM-R: An chemical reasoner LLM enhanced with atomized chemical knowledge},
  author={Zhao, Zihan and Chen, Bo and Wan, Ziping and Chen, Lu and Lin, Xuanze and Yu, Shiyang and Zhang, Situo and Ma, Da and Zhu, Zichen and Zhang, Danyang and others},
  journal={arXiv preprint arXiv:2507.21990},
  year={2025}
}

@article{jablonka2024llmchem,
  title={Leveraging large language models for predictive chemistry},
  author={Jablonka, Kevin Maik and Schwaller, Philippe and Ortega-Guerrero, Andres and Smit, Berend},
  journal={Nature Machine Intelligence},
  volume={6},
  number={2},
  pages={161--169},
  year={2024},
  publisher={Nature Publishing Group UK London}
}

@article{wei2022cot,
  title={Chain-of-thought prompting elicits reasoning in large language models},
  author={Wei, Jason and Wang, Xuezhi and Schuurmans, Dale and Bosma, Maarten and Xia, Fei and Chi, Ed and Le, Quoc V and Zhou, Denny and others},
  journal={Advances in neural information processing systems},
  volume={35},
  pages={24824--24837},
  year={2022}
}

@article{tzschucke2002workup,
  title={Modern separation techniques for the efficient workup in organic synthesis},
  author={Tzschucke, Carl Christoph and Markert, Christian and Bannwarth, Willi and Roller, Sebastian and Hebel, Andr{\'e} and Haag, Rainer},
  journal={Angewandte Chemie International Edition},
  volume={41},
  number={21},
  pages={3964--4000},
  year={2002},
  publisher={Wiley Online Library}
}

@article{matous2021workup,
  title={Reaction Outcome Critically Dependent on the Method of Workup: An Example from the Synthesis of 1-Isoquinolones},
  author={Matous, Petr and Majek, Michal and Kysilka, Ondrej and Kunes, Jiri and Marikova, Jana and Ruzicka, Ales and Pour, Milan and Kocovsky, Pavel},
  journal={The Journal of Organic Chemistry},
  volume={86},
  number={12},
  pages={8078--8088},
  year={2021},
  publisher={ACS Publications}
}

@article{zhang2025chemma,
  title={Large language models to accelerate organic chemistry synthesis},
  author={Zhang, Yu and Han, Yang and Chen, Shuai and Yu, Ruijie and Zhao, Xin and Liu, Xianbin and Zeng, Kaipeng and Yu, Mengdi and Tian, Jidong and Zhu, Feng and others},
  journal={Nature Machine Intelligence},
  pages={1--13},
  year={2025},
  publisher={Nature Publishing Group UK London}
}

@article{wang2025chemR,
  title={Chem-R: Learning to Reason as a Chemist},
  author={Wang, Weida and Chen, Benteng and Zhang, Di and Liu, Wanhao and Pu, Shuchen and Gao, Ben and Zeng, Jin and Bai, Lei and Ouyang, Wanli and Wei, Xiaoyong and others},
  journal={arXiv preprint arXiv:2510.16880},
  year={2025}
}

@article{vaswani2017transformer,
  title={Attention is all you need},
  author={Vaswani, Ashish and Shazeer, Noam and Parmar, Niki and Uszkoreit, Jakob and Jones, Llion and Gomez, Aidan N and Kaiser, {\L}ukasz and Polosukhin, Illia},
  journal={Advances in neural information processing systems},
  volume={30},
  year={2017}
}

@article{lewis2019bart,
  title={BART: Denoising sequence-to-sequence pre-training for natural language generation, translation, and comprehension},
  author={Lewis, Mike and Liu, Yinhan and Goyal, Naman and Ghazvininejad, Marjan and Mohamed, Abdelrahman and Levy, Omer and Stoyanov, Ves and Zettlemoyer, Luke},
  journal={arXiv preprint arXiv:1910.13461},
  year={2019}
}

@article{lindahl2025pronounced,
  title={Pronounced electronic modulation of geometrically-regulated metalloenediyne cyclization},
  author={Lindahl, Sarah E and Metzger, Erin M and Chen, Chun-Hsing and Pink, Maren and Zaleski, Jeffrey M},
  journal={Chemical Science},
  volume={16},
  number={1},
  pages={255--279},
  year={2025},
  publisher={Royal Society of Chemistry}
}

@misc{stockley2020autotaxinpatent,
  title={Autotaxin inhibitory compounds},
  author={Stockley, Martin Lee and MACDONALD, Ellen Catherine and Shah, Pritom and Jordan, Allan and Hitchin, James and Hamilton, Niall},
  year={2020},
  month=may # "~19",
  publisher={Google Patents},
  note={US Patent 10,654,846}
}

@article{albeiicio2001peptide,
  title={New trends in peptide coupling reagents},
  author={Albeiicio, Fernando and Chinchilla, Rafael and Dodsworth, David J and Najera, Carmen},
  journal={Organic Preparations and Procedures International},
  volume={33},
  number={3},
  pages={203--303},
  year={2001},
  publisher={Taylor \& Francis}
}

@article{woodward1954strychnine,
  title={The total synthesis of strychnine},
  author={Woodward, Robert B and Cava, Michael P and Ollis, William David and Hunger, A and Daeniker, HU and Schenker, K},
  journal={Journal of the American Chemical Society},
  volume={76},
  number={18},
  pages={4749--4751},
  year={1954},
  publisher={ACS Publications}
}

@article{sheng2024hybridflow_verl,
  title   = {HybridFlow: A Flexible and Efficient RLHF Framework},
  author  = {Guangming Sheng and Chi Zhang and Zilingfeng Ye and Xibin Wu and Wang Zhang and Ru Zhang and Yanghua Peng and Haibin Lin and Chuan Wu},
  year    = {2024},
  journal = {arXiv preprint arXiv: 2409.19256}
}

@article{vavskevivcius2024language,
  title={Language models for predicting organic synthesis procedures},
  author={Va{\v{s}}kevi{\v{c}}ius, Mantas and Kapo{\v{c}}i{\=u}t{\.e}-Dzikien{\.e}, Jurgita},
  journal={Applied Sciences},
  volume={14},
  number={24},
  pages={11526},
  year={2024},
  publisher={MDPI}
}

@article{shim2025recommending,
  title={Recommending reaction conditions with label ranking},
  author={Shim, Eunjae and Tewari, Ambuj and Cernak, Tim and Zimmerman, Paul M},
  journal={Chemical Science},
  volume={16},
  number={9},
  pages={4109--4118},
  year={2025},
  publisher={Royal Society of Chemistry}
}

@article{tu2023predictive,
  title={Predictive chemistry: machine learning for reaction deployment, reaction development, and reaction discovery},
  author={Tu, Zhengkai and Stuyver, Thijs and Coley, Connor W},
  journal={Chemical science},
  volume={14},
  number={2},
  pages={226--244},
  year={2023},
  publisher={Royal Society of Chemistry}
}

@article{ouyang2022training,
  title={Training language models to follow instructions with human feedback},
  author={Ouyang, Long and Wu, Jeffrey and Jiang, Xu and Almeida, Diogo and Wainwright, Carroll and Mishkin, Pamela and Zhang, Chong and Agarwal, Sandhini and Slama, Katarina and Ray, Alex and others},
  journal={Advances in neural information processing systems},
  volume={35},
  pages={27730--27744},
  year={2022}
}

@article{m2024augmenting,
  title={Augmenting large language models with chemistry tools},
  author={M. Bran, Andres and Cox, Sam and Schilter, Oliver and Baldassari, Carlo and White, Andrew D and Schwaller, Philippe},
  journal={Nature Machine Intelligence},
  volume={6},
  number={5},
  pages={525--535},
  year={2024},
  publisher={Nature Publishing Group UK London}
}

@article{xia2025naturelm,
  title={Nature Language Model: Deciphering the Language of Nature for Scientific Discovery},
  author={Xia, Yingce and Jin, Peiran and Xie, Shufang and He, Liang and Cao, Chuan and Luo, Renqian and Liu, Guoqing and Wang, Yue and Liu, Zequn and Chen, Yuan-Jyue and others},
  journal={arXiv preprint arXiv:2502.07527},
  year={2025}
}

@article{ether0,
  title={Training a Scientific Reasoning Model for Chemistry},
  author={Narayanan, Siddharth M and Braza, James D and Griffiths, Ryan-Rhys and Bou, Albert and Wellawatte, Geemi and Ramos, Mayk Caldas and Mitchener, Ludovico and Rodriques, Samuel G and White, Andrew D},
  journal={arXiv preprint arXiv:2506.17238},
  year={2025}
}

@article{guo2025deepseek,
  title={Deepseek-r1 incentivizes reasoning in llms through reinforcement learning},
  author={Guo, Daya and Yang, Dejian and Zhang, Haowei and Song, Junxiao and Wang, Peiyi and Zhu, Qihao and Xu, Runxin and Zhang, Ruoyu and Ma, Shirong and Bi, Xiao and others},
  journal={Nature},
  volume={645},
  number={8081},
  pages={633--638},
  year={2025},
  publisher={Nature Publishing Group UK London}
}

@article{liu2017retrosynthetic,
  title={Retrosynthetic reaction prediction using neural sequence-to-sequence models},
  author={Liu, Bowen and Ramsundar, Bharath and Kawthekar, Prasad and Shi, Jade and Gomes, Joseph and Luu Nguyen, Quang and Ho, Stephen and Sloane, Jack and Wender, Paul and Pande, Vijay},
  journal={ACS central science},
  volume={3},
  number={10},
  pages={1103--1113},
  year={2017},
  publisher={ACS Publications}
}

@article{schwaller2019molecular,
  title={Molecular transformer: a model for uncertainty-calibrated chemical reaction prediction},
  author={Schwaller, Philippe and Laino, Teodoro and Gaudin, Th{\'e}ophile and Bolgar, Peter and Hunter, Christopher A and Bekas, Costas and Lee, Alpha A},
  journal={ACS central science},
  volume={5},
  number={9},
  pages={1572--1583},
  year={2019},
  publisher={ACS Publications}
}

@article{boiko2023autonomous,
  title={Autonomous chemical research with large language models},
  author={Boiko, Daniil A and MacKnight, Robert and Kline, Ben and Gomes, Gabe},
  journal={Nature},
  volume={624},
  number={7992},
  pages={570--578},
  year={2023},
  publisher={Nature Publishing Group UK London}
}

@article{pace20122MeTHF,
  title={2-Methyltetrahydrofuran (2-MeTHF): a biomass-derived solvent with broad application in organic chemistry},
  author={Pace, Vittorio and Hoyos, Pilar and Castoldi, Laura and Dom{\'\i}nguez de Mar{\'\i}a, Pablo and Alc{\'a}ntara, Andr{\'e}s R},
  journal={ChemSusChem},
  volume={5},
  number={8},
  pages={1369--1379},
  year={2012},
  publisher={Wiley Online Library}
}

@article{clark2015machines,
  title={Machines first, humans second: on the importance of algorithmic interpretation of open chemistry data},
  author={Clark, Alex M and Williams, Antony J and Ekins, Sean},
  journal={Journal of cheminformatics},
  volume={7},
  number={1},
  pages={9},
  year={2015},
  publisher={Springer}
}

@article{davies2019digitization,
  title={The digitization of organic synthesis},
  author={Davies, Ian W},
  journal={Nature},
  volume={570},
  number={7760},
  pages={175--181},
  year={2019},
  publisher={Nature Publishing Group UK London}
}

@article{steiner2019organicautomatic,
  title={Organic synthesis in a modular robotic system driven by a chemical programming language},
  author={Steiner, Sebastian and Wolf, Jakob and Glatzel, Stefan and Andreou, Anna and Granda, Jaros{\l}aw M and Keenan, Graham and Hinkley, Trevor and Aragon-Camarasa, Gerardo and Kitson, Philip J and Angelone, Davide and others},
  journal={Science},
  volume={363},
  number={6423},
  pages={eaav2211},
  year={2019},
  publisher={American Association for the Advancement of Science}
}

@article{coley2019robotic,
  title={A robotic platform for flow synthesis of organic compounds informed by AI planning},
  author={Coley, Connor W and Thomas III, Dale A and Lummiss, Justin AM and Jaworski, Jonathan N and Breen, Christopher P and Schultz, Victor and Hart, Travis and Fishman, Joshua S and Rogers, Luke and Gao, Hanyu and others},
  journal={Science},
  volume={365},
  number={6453},
  pages={eaax1566},
  year={2019},
  publisher={American Association for the Advancement of Science}
}

@article{mehr2020automaticchem,
  title={A universal system for digitization and automatic execution of the chemical synthesis literature},
  author={Mehr, S Hessam M and Craven, Matthew and Leonov, Artem I and Keenan, Graham and Cronin, Leroy},
  journal={Science},
  volume={370},
  number={6512},
  pages={101--108},
  year={2020},
  publisher={American Association for the Advancement of Science}
}

@article{schulman2017proximal,
  title={Proximal policy optimization algorithms},
  author={Schulman, John and Wolski, Filip and Dhariwal, Prafulla and Radford, Alec and Klimov, Oleg},
  journal={arXiv preprint arXiv:1707.06347},
  year={2017}
}

@article{zeng2024token,
  title={Token-level direct preference optimization},
  author={Zeng, Yongcheng and Liu, Guoqing and Ma, Weiyu and Yang, Ning and Zhang, Haifeng and Wang, Jun},
  journal={arXiv preprint arXiv:2404.11999},
  year={2024}
}

@article{wang2025reinforcement,
  title={Reinforcement learning for reasoning in large language models with one training example},
  author={Wang, Yiping and Yang, Qing and Zeng, Zhiyuan and Ren, Liliang and Liu, Liyuan and Peng, Baolin and Cheng, Hao and He, Xuehai and Wang, Kuan and Gao, Jianfeng and others},
  journal={arXiv preprint arXiv:2504.20571},
  year={2025}
}

@article{hurst2024gpt4o,
  title={Gpt-4o system card},
  author={Hurst, Aaron and Lerer, Adam and Goucher, Adam P and Perelman, Adam and Ramesh, Aditya and Clark, Aidan and Ostrow, AJ and Welihinda, Akila and Hayes, Alan and Radford, Alec and others},
  journal={arXiv preprint arXiv:2410.21276},
  year={2024}
}

@article{shao2024deepseekmath,
  title={Deepseekmath: Pushing the limits of mathematical reasoning in open language models},
  author={Shao, Zhihong and Wang, Peiyi and Zhu, Qihao and Xu, Runxin and Song, Junxiao and Bi, Xiao and Zhang, Haowei and Zhang, Mingchuan and Li, YK and Wu, Yang and others},
  journal={arXiv preprint arXiv:2402.03300},
  year={2024}
}

@article{schulman2015high,
  title={High-dimensional continuous control using generalized advantage estimation},
  author={Schulman, John and Moritz, Philipp and Levine, Sergey and Jordan, Michael and Abbeel, Pieter},
  journal={arXiv preprint arXiv:1506.02438},
  year={2015}
}

@article{mirza2025framework,
  title={A framework for evaluating the chemical knowledge and reasoning abilities of large language models against the expertise of chemists},
  author={Mirza, Adrian and Alampara, Nawaf and Kunchapu, Sreekanth and R{\'\i}os-Garc{\'\i}a, Marti{\~n}o and Emoekabu, Benedict and Krishnan, Aswanth and Gupta, Tanya and Schilling-Wilhelmi, Mara and Okereke, Macjonathan and Aneesh, Anagha and others},
  journal={Nature Chemistry},
  pages={1--8},
  year={2025},
  publisher={Nature Publishing Group UK London}
}

@article{jaech2024o1,
  title={Openai o1 system card},
  author={Jaech, Aaron and Kalai, Adam and Lerer, Adam and Richardson, Adam and El-Kishky, Ahmed and Low, Aiden and Helyar, Alec and Madry, Aleksander and Beutel, Alex and Carney, Alex and others},
  journal={arXiv preprint arXiv:2412.16720},
  year={2024}
}

@article{bran2025chemical,
  title={Chemical reasoning in LLMs unlocks steerable synthesis planning and reaction mechanism elucidation},
  author={Bran, Andres M and Neukomm, Theo A and Armstrong, Daniel P and Jon{\v{c}}ev, Zlatko and Schwaller, Philippe},
  journal={arXiv preprint arXiv:2503.08537},
  year={2025}
}

@article{li2025ChemCoTBench,
  title={Beyond Chemical QA: Evaluating LLM's Chemical Reasoning with Modular Chemical Operations},
  author={Li, Hao and Cao, He and Feng, Bin and Shao, Yanjun and Tang, Xiangru and Yan, Zhiyuan and Yuan, Li and Tian, Yonghong and Li, Yu},
  journal={arXiv preprint arXiv:2505.21318},
  year={2025}
}

@misc{qwen3max,
    title = {Qwen3-Max: Just Scale it},
    author = {Qwen Team},
    month = {September},
    year = {2025}
}

@article{campos2019importance,
  title={The importance of synthetic chemistry in the pharmaceutical industry},
  author={Campos, Kevin R and Coleman, Paul J and Alvarez, Juan C and Dreher, Spencer D and Garbaccio, Robert M and Terrett, Nicholas K and Tillyer, Richard D and Truppo, Matthew D and Parmee, Emma R},
  journal={Science},
  volume={363},
  number={6424},
  pages={eaat0805},
  year={2019},
  publisher={American Association for the Advancement of Science}
}

@article{strieth2020machine,
  title={Machine learning the ropes: principles, applications and directions in synthetic chemistry},
  author={Strieth-Kalthoff, Felix and Sandfort, Frederik and Segler, Marwin HS and Glorius, Frank},
  journal={Chemical Society Reviews},
  volume={49},
  number={17},
  pages={6154--6168},
  year={2020},
  publisher={Royal Society of Chemistry}
}

@article{ai2024extracting,
  title={Extracting structured data from organic synthesis procedures using a fine-tuned large language model},
  author={Ai, Qianxiang and Meng, Fanwang and Shi, Jiale and Pelkie, Brenden and Coley, Connor W},
  journal={Digital discovery},
  volume={3},
  number={9},
  pages={1822--1831},
  year={2024},
  publisher={Royal Society of Chemistry}
}

@article{zhao2025literature,
  title={From literature to lab protocols with knowledge-graph-guided large language models},
  author={Zhao, Boyao and Li, Zhengwen and Zhang, Yihan and Song, Tao and Shi, Zhou and Lin, Quan and Shang, Weiwei and Jiang, Jun and Chen, Linjiang},
  year={2025}
}

@article{mendes2025automated,
  title={Automated LLM based Extraction of Standardized Synthesis Procedures: an All-Domain, Zero-Shot Approach},
  author={Mendes, Pedro and Costa, Daniel and Manica, Matteo and Laino, Teodoro and Ribeiro, Filipa},
  year={2025}
}

@article{liu2025fgbench,
  title={FGBench: A Dataset and Benchmark for Molecular Property Reasoning at Functional Group-Level in Large Language Models},
  author={Liu, Xuan and Ouyang, Siru and Zhong, Xianrui and Han, Jiawei and Zhao, Huimin},
  journal={arXiv preprint arXiv:2508.01055},
  year={2025}
}

@article{machi2025actionsurvey,
  title={A framework for reviewing the results of automated conversion of structured organic synthesis procedures from the literature},
  author={Machi, Kojiro and Akiyama, Seiji and Nagata, Yuuya and Yoshioka, Masaharu},
  journal={Digital Discovery},
  volume={4},
  number={1},
  pages={172--180},
  year={2025},
  publisher={Royal Society of Chemistry}
}

@inproceedings{yuan2025uspto,
  title={USPTO-LLM: A Large Language Model-Assisted Information-enriched Chemical Reaction Dataset},
  author={Yuan, Shen and Gong, Shukai and Xu, Hongteng},
  booktitle={Companion Proceedings of the ACM on Web Conference 2025},
  pages={817--820},
  year={2025}
}

@article{zhang2025chemactor,
  title={ChemActor: Enhancing Automated Extraction of Chemical Synthesis Actions with LLM-Generated Data},
  author={Zhang, Yu and Yu, Ruijie and Tian, Jidong and Zhu, Feng and Liu, Jiapeng and Yang, Xiaokang and Jin, Yaohui and Xu, Yanyan},
  journal={arXiv preprint arXiv:2506.23520},
  year={2025}
}

@misc{zhao2025superchem,
      title={SUPERChem: A Multimodal Reasoning Benchmark in Chemistry}, 
      author={Zehua Zhao and Zhixian Huang and Junren Li and Siyu Lin and Junting Zhou and Fengqi Cao and Kun Zhou and Rui Ge and Tingting Long and Yuexiang Zhu and Yan Liu and Jie Zheng and Junnian Wei and Rong Zhu and Peng Zou and Wenyu Li and Zekai Cheng and Tian Ding and Yaxuan Wang and Yizhao Yan and Tingru Wei and Haowei Ming and Weijie Mao and Chen Sun and Yiming Liu and Zichen Wang and Zuo Zhang and Tong Yang and Hao Ma and Zhen Gao and Jian Pei},
      year={2025},
      eprint={2512.01274},
      archivePrefix={arXiv},
      primaryClass={cs.CL},
      url={https://arxiv.org/abs/2512.01274}, 
}

@article{zhao2025molreasoner,
  title={Molreasoner: Toward effective and interpretable reasoning for molecular llms},
  author={Zhao, Guojiang and Li, Sihang and Lu, Zixiang and Cheng, Zheng and Lin, Haitao and Wu, Lirong and Xia, Hanchen and Cai, Hengxing and Guo, Wentao and Wang, Hongshuai and others},
  journal={arXiv preprint arXiv:2508.02066},
  year={2025}
}

@article{vleduts1963concerning,
  title={Concerning one system of classification and codification of organic reactions},
  author={Vleduts, GE},
  journal={Information Storage and Retrieval},
  volume={1},
  number={2-3},
  pages={117--146},
  year={1963},
  publisher={Elsevier}
}

@article{corey1969computer,
  title={Computer-Assisted Design of Complex Organic Syntheses: Pathways for molecular synthesis can be devised with a computer and equipment for graphical communication.},
  author={Corey, Elias James and Wipke, W Todd},
  journal={Science},
  volume={166},
  number={3902},
  pages={178--192},
  year={1969},
  publisher={American Association for the Advancement of Science}
}

@article{dai2019retrosynthesis,
  title={Retrosynthesis prediction with conditional graph logic network},
  author={Dai, Hanjun and Li, Chengtao and Coley, Connor and Dai, Bo and Song, Le},
  journal={Advances in Neural Information Processing Systems},
  volume={32},
  year={2019}
}

@article{chen2021deep,
  title={Deep retrosynthetic reaction prediction using local reactivity and global attention},
  author={Chen, Shuan and Jung, Yousung},
  journal={JACS Au},
  volume={1},
  number={10},
  pages={1612--1620},
  year={2021},
  publisher={ACS Publications}
}

@inproceedings{xie2023retrosynthesis,
  title={Retrosynthesis prediction with local template retrieval},
  author={Xie, Shufang and Yan, Rui and Guo, Junliang and Xia, Yingce and Wu, Lijun and Qin, Tao},
  booktitle={Proceedings of the AAAI Conference on Artificial Intelligence},
  volume={37},
  number={4},
  pages={5330--5338},
  year={2023}
}

@article{gainski2025diverse,
  title={Diverse and feasible retrosynthesis using GFlowNets},
  author={Gai{\'n}ski, Piotr and Koziarski, Micha{\l} and Maziarz, Krzysztof and Segler, Marwin and Tabor, Jacek and {\'S}mieja, Marek},
  journal={Information Sciences},
  volume={714},
  pages={122194},
  year={2025},
  publisher={Elsevier}
}

@inproceedings{xie2022retrograph,
  title={Retrograph: Retrosynthetic planning with graph search},
  author={Xie, Shufang and Yan, Rui and Han, Peng and Xia, Yingce and Wu, Lijun and Guo, Chenjuan and Yang, Bin and Qin, Tao},
  booktitle={Proceedings of the 28th ACM SIGKDD Conference on Knowledge Discovery and Data Mining},
  pages={2120--2129},
  year={2022}
}

@article{tripp2023retro,
  title={Retro-fallback: retrosynthetic planning in an uncertain world},
  author={Tripp, Austin and Maziarz, Krzysztof and Lewis, Sarah and Segler, Marwin and Hern{\'a}ndez-Lobato, Jos{\'e} Miguel},
  journal={arXiv preprint arXiv:2310.09270},
  year={2023}
}

@article{comanici2025gemini,
  title={Gemini 2.5: Pushing the frontier with advanced reasoning, multimodality, long context, and next generation agentic capabilities},
  author={Comanici, Gheorghe and Bieber, Eric and Schaekermann, Mike and Pasupat, Ice and Sachdeva, Noveen and Dhillon, Inderjit and Blistein, Marcel and Ram, Ori and Zhang, Dan and Rosen, Evan and others},
  journal={arXiv preprint arXiv:2507.06261},
  year={2025}
}

@misc{anthropic2024sonnet35,
  title = {Anthropic Introducing Claude 3.5 Sonnet.},
  howpublished = {\url{https://www.anthropic.com/news/
 claude-3-5-sonnet}},
  year={2024},
  note={Press release announcing Claude 3.5 Sonnet with benchmark results.}
}

@misc{pistachio,
  author = {NextMove},
  title = {Pistachio},
  howpublished = {\url{http://www.nextmovesoftware.com/pistachio.html}},
}

@article{YAMAMOTO2025,
title = {Lessons Learned during 50 kg Manufacturing of Suzuki–Miyaura Coupling Reaction},
journal = {Organic Process Research \& Development},
year = {2025},
issn = {1083-6160},
doi = {https://doi.org/10.1021/acs.oprd.5c00207},
url = {https://www.sciencedirect.com/science/article/pii/S1083616025001781},
author = {Yuhei Yamamoto and Kotaro Yamaguchi and Kentaro Yaji},
}

@article{abdin2025phi,
  title={Phi-4-reasoning technical report},
  author={Abdin, Marah and Agarwal, Sahaj and Awadallah, Ahmed and Balachandran, Vidhisha and Behl, Harkirat and Chen, Lingjiao and de Rosa, Gustavo and Gunasekar, Suriya and Javaheripi, Mojan and Joshi, Neel and others},
  journal={arXiv preprint arXiv:2504.21318},
  year={2025}
}

\clearpage

\section*{Appendix}

\subsection*{1. Large-Scale Procedure Dataset Construction via LLM Annotation}

\subsubsection*{1.1. Action system definitions}\label{app:action_system}

To convert free‑form experimental descriptions into structured procedures, we defined an action system that captures the most common operations used in routine chemical reaction experiments under typical laboratory setups. The complete set of action types is summarized in Table~\ref{tab:action_types_combined}.

\begin{table*}
\centering  
\begin{tabular}{ll}  
\toprule  
\textbf{Action Name} & \textbf{Description / Inputs \& Outputs} \\  
\midrule 
\textbf{Add} & Add source substances to target chemicals or reaction mixtures. \\  
 & \textbf{Inputs: source, target, time period, method}, \textbf{Outputs: Mixture} \\  
\textbf{Change atmosphere} & Set atmosphere of target substances or reaction mixtures. \\  
 & \textbf{Inputs: target, atmosphere}, \textbf{Outputs: None} \\  
\textbf{Change pH} & Change pH of target substances or reaction mixtures. \\  
 & \textbf{Inputs: target, pH, agent}, \textbf{Outputs: None} \\
\textbf{Change pressure} & Change pressure of target substances or reaction mixtures. \\  
 & \textbf{Inputs: target, pressure, apparatus}, \textbf{Outputs: None} \\  
\textbf{Change temperature} & Change temperature of target substances or reaction mixtures. \\  
 & \textbf{Inputs}: target, temperature, speed, apparatus, agent. \textbf{Outputs: None} \\ 
\textbf{Chromatograph} & Purify reaction mixtures by passing them through a chromatography column. \\  
 & \textbf{Inputs: target, column, eluent}, \textbf{Outputs: Mixture} \\
\textbf{Concentrate} & Remove solvents to concentrate reaction mixtures. \\  
 & \textbf{Inputs: target, in vacuum, apparatus}, \textbf{Outputs: Mixture} \\
\textbf{Degas} & Purge target substances or reaction mixtures with a gas.
\\  
 & \textbf{Inputs: target, agent, time period}, \textbf{Outputs: None} \\  
\textbf{Distill} & Distill reaction mixtures to remove agents. \\  
 & \textbf{Inputs: target, agent to remove, apparatus}, \textbf{Outputs: Mixture} \\  
\textbf{Dry} & Remove residual solvents from reaction mixtures using an agent or vacuum. \\  
 & \textbf{Inputs: target, in vacuum, agent, apparatus}, \textbf{Outputs: Mixture} \\  
\textbf{Extract} & Transfer compounds from one phase to another using an different solvent. \\  
 & \textbf{Inputs: target, agent, times}, \textbf{Outputs: Mixture} \\ 
\textbf{Filter solution} & Separates solid and liquid phases using a filtration apparatus. \\  
 & \textbf{Inputs: target, apparatus}, \textbf{Outputs: Mixture, Mixture} \\  
\textbf{Irradiate} & Use controlled light exposure on target substances or reaction mixtures. \\  
 & \textbf{Inputs: target, time period, apparatus, wavelength}, \textbf{Outputs: None} \\  
\textbf{Make solution} & Dissolve solutes in solvents to obtain a mixture. \\  
 & \textbf{Inputs: solute, solvent, container}, \textbf{Outputs: Mixture} \\ 
\textbf{Microwave} & Heat target substances or reaction mixtures in a microwave apparatus.\\  
 & \textbf{Inputs: target, time period, apparatus}, \textbf{Outputs: None} \\ 
\textbf{Other purification} & Purify reaction mixtures using other methods. \\  
 & \textbf{Inputs: target, method, agent, apparatus}, \textbf{Outputs: Mixture} \\  
\textbf{Partition} & Separate reaction mixtures into layers via two immiscible solvents. \\  
 & \textbf{Inputs: target, solvents 1, solvents 2}, \textbf{Outputs: Mixture, Mixture} \\  
\textbf{Quench} & Stop reaction by adding a substance. \\  
 & \textbf{Inputs: target, agent}, \textbf{Outputs: Mixture} \\
\textbf{Recrystallize} & Recrystallize solid reaction mixtures from solvents. \\  
 & \textbf{Inputs: target, solvent, times}, \textbf{Outputs: Mixture} \\ 
\textbf{Sample} & Take a quantity from source chemicals or reaction mixtures. \\  
 & \textbf{Inputs: source, quantity}, \textbf{Outputs: Chemical/Mixture} \\  
\textbf{Sonicate} & Agitate solutions with sound waves. \\  
 & \textbf{Inputs: target, time period, apparatus}, \textbf{Outputs: None} \\ 
\textbf{Triturate} & Triturate reaction mixtures under conditions. \\  
 & \textbf{Inputs: target, condition, apparatus}, \textbf{Outputs: Mixture} \\  
\textbf{Wait} & Leave reaction mixtures to stand for a specified duration. \\  
 & \textbf{Inputs: time period}, \textbf{Outputs: None} \\  
\textbf{Wash} & Wash reaction mixtures with solvents. \\  
 & \textbf{Inputs: target, solvent, times}, \textbf{Outputs: Mixture} \\  
\textbf{Yield statement} & Record yield information of obtained products. \\  
 & \textbf{Inputs: product, target, yield, quantities, purity}, \textbf{Outputs: None} \\  
\bottomrule
\end{tabular}  
\caption{Action types for describing experimental procedures.}  
\label{tab:action_types_combined}  
\end{table*}

\subsubsection*{1.2. Prompts used in LLM annotation}

The automatic action annotation pipeline comprises three main stages: (1) Coreference resolution, (2) Code generation and execution, and (3) Verification. The detailed prompt templates for each stage are provided in Figure~\ref{fig:cr_prompt}, Figure~\ref{fig:code_gen_prompt}, and Figure~\ref{fig:verify_prompt}.

\begin{figure*}[htbp]
\begin{myminted}{Coreference Resolution Template}
You are an expert in natural language processing and chemistry, specifically chemical mentions identification and coreference resolution. Given a text passage of experimental operations and a list of chemicals, your task is to first identify other chemicals that appear in the text passage, then identify all mentions of these entities and replace each mention with the entity's corresponding ID enclosed in `\$'.\\

Please keep the original sentence structure and grammar as much as possible (meaning that the text passage could be restored to its original form by replacing the IDs back with the corresponding mentions), unless there are errors in the original text such as typos or misplacements.\\

Note that ice and water should not be considered as the same chemical due to the difference in temperature.\\

Please STRICTLY follow the format of the provided example and the instructions. No other headings and information should be added.\\

Instructions:

1. Read the provided text passage carefully.\\
2. Consider the list of entities and their possible names/mentions. This list provides the ground truth for coreference.\\
3. Identify all other chemicals that appear in the text passage.\\
4. Extend the list of IDs and corresponding coreference of the chemicals.\\
5. Identify all mentions of these entities within the text passage. Note that not all strings that match the coreference are entity mentions. Some of them may serve other purposes and have other meanings.\\
4. Replace each identified mention with the entity's ID enclosed in `\$'. Do not forget the `\$' and be sure that the ID is from the extended list of entities and corresponding to the correct entity.\\
5. If a mention refers to an entity NOT in the provided list, leave it as is. Do not create new entity IDs.\\
6. Preserve the original sentence structure and grammar of the text passage as much as possible. Only replace the mentions with IDs.

Example:\\
\textcolor{red}{\{Examples\}}\\

Now, apply these instructions to the following text passage and entities:

Text Passage:\\
\textcolor{red}{\{Paragraphtext\}}\\

Entities and Possible Mentions:\\
\textcolor{red}{\{Mapping\}}\\

Other Chemicals Mentioned:
\end{myminted}
\caption{Coreference resolution prompt used in LLM-based procedure annotation.}
\label{fig:cr_prompt}
\end{figure*}

\begin{figure*}[htbp]
\begin{myminted}{Code Generation Template}

You are provided with a textual description of a chemistry experiment, along with an outline of Python functions that represent various chemical operations. Your task is to accurately translate the described experiment into a Python script using these functions. The translation must be faithful to the original description, ensuring that all details are included and no information is omitted or contradicted. You may enhance the code with additional context if it aids clarity, but you must not introduce any inaccuracies.\\

Please note that within the provided text, chemicals may be enclosed in `\$'. These markings are guidelines and might contain errors or omissions, so use your best judgment based on the context to identify the correct chemicals.\\

Functions:\\
\textcolor{red}{\{Functions\}}\\

Requirements:\\
1. Please double-check that the generated code does not have any unexpected keyword arguments that are not present in the provided function definitions.\\
2. The function \texttt{supplement\_information} should be used as less as possible and only when necessary, try to avoid using it by utilizing other functions to achieve the same result.\\
3. Don't assume any pre-defined variables; define all necessary variables within the code.\\
4. The last line of the code MUST be the function  \texttt{yield\_statement}, which provides the final output of the experiment.\\
5. Please ONLY output the python code enclosed in \texttt{"""},  
starting with \texttt{"""python} and ending with \texttt{"""}.\\

Examples:\\
\textcolor{red}{\{Examples\}}\\

Now, convert the following paragraph into Python code.\\

Description:\\
\textcolor{red}{\{Paragraphtext\}}\\

Code:

\end{myminted}
\caption{Code generation prompt used in LLM-based procedure annotation.}
\label{fig:code_gen_prompt}
\end{figure*}

\begin{figure*}
\begin{myminted}{Verification Template}

You are tasked with analyzing two text paragraphs to determine if the query paragraph describe the same chemical experiment process as the reference paragraph. The core information in both paragraphs must be exactly the same, including the substances used, their quantities, and the procedures performed. While it is permissible for some information to be omitted or supplemented reasonably—such as missing one of multiple measurement methods—it is not permissible for there to be conflicting information or the omission of key information, such as a missing substance.\\

Please perform a thorough analysis and reasoning process, which should be enclosed within <Think> and </Think>. \\
Following your analysis, output <Answer>Yes</Answer> if the paragraphs describe the same experiment process, <Answer>No</Answer> if they do not, or <Answer>Uncertain</Answer> if you are unsure.\\
Lastly, provide a confidence level from 1 to 5, wrapped in <Confidence></Confidence>, where 1 is the lowest and 5 is the highest.\\

Example Output Format:\\
<Think>[Your thorough analysis and reasoning here]</Think>\\
<Answer>[Yes/No/Uncertain]</Answer>\\
<Confidence>[1-5]</Confidence>\\

Please STRICTLY follow the output format above.\\
<Reference>
\textcolor{red}{\{Reference\_Paragraphtext\}}
</Reference>\\

<Query>
\textcolor{red}{\{Query\_Paragraphtext\}}
</Query>

\end{myminted}
\caption{Verification prompt used in LLM-based procedure annotation.}
\label{fig:verify_prompt}
\end{figure*}

\subsubsection*{1.3. Comparative analysis of structured procedures: Our action system vs. OpenExp}

To ensure a fair comparison between our action system and those employed in OpenExp~\cite{liu2024reactxt}, we first downloaded the OpenExp dataset\footnote{\href{https://github.com/syr-cn/ReactXT/tree/master/openExp}{\texttt{https://github.com/syr-cn/ReactXT/openExp}}}. We then applied our LLM-based automated annotation pipeline to these reactions, producing structured action sequences that follow our action schema. Finally, we used the \texttt{o3-high} model as an LLM-as-a-judge to evaluate a subset of approximately 80,000 entries (constrained by API throughput).
  
The evaluation prompt is as follows:  
  
\begin{quote}  
``I will give you a chemical reaction and two corresponding transcribed action sequences. Please determine whether each sequence is consistent with the chemical reaction, and assign an overall score for each reaction according to the following scheme (0--10): 10 = fully consistent; 8--9 = only minor, harmless deviations; 5--7 = noticeable deviations but the sequence would still yield essentially the same result; 1--4 = important steps are wrong or missing; 0 = not related to the procedure. In addition to the overall score, please provide scores in three categories: substance score (correctness of substances), action score (coverage and correctness of actions), and order score (alignment of action sequence with the original text). Begin with a brief analysis, then report scores in the format: ``Action 1: Overall x/10; Substance score x/10; Action score x/10; Order score x/10. Action 2: Overall x/10; Substance score x/10; Action score x/10; Order score x/10.''  
\end{quote}  
  
The textual experimental description and its corresponding action sequences were inserted into the prompt in the following format:
  
\begin{quote}
``Reaction description: \{paragraphtext\},\\  
Action series 1: \{action\_1\},\\  
Action series 2: \{action\_2\}.''
\end{quote}

\subsubsection*{1.4. Statistics on preprocessing the raw Pistachio dataset}

The raw Pistachio dataset (2024Q2 version, \texttt{US-grants} folder) contains 4,410,491 entries.   
We first applied a series of cleaning steps:  
  
\begin{enumerate}  
    \item Removed reactions with invalid SMILES representations (as detected by RDKit).  
    \item Removed reactions in which reactants or products did not contain any carbon atoms (\texttt{c} or \texttt{C}).  
    \item Removed reactions with potential atom-mapping errors.  
    \item Deduplicated reactions based on reaction SMILES without atom mapping.  
\end{enumerate}  
  
After cleaning, the dataset contained 2,061,352 reactions. During the LLM-based automated annotation stage:
  
\begin{itemize}  
    \item Coreference resolution step: Excluded reactions whose paragraphText contained abandoned keywords (e.g., ``as described for example''), consisted of fewer than two sentences, or whose restored paragraph had an edit distance from the original paragraphText exceeding a predefined threshold.  
    \item Code generation step: Removed reactions if the generated code was invalid (e.g., failed Python syntax checks, lacked a \texttt{yield\_statement}), or if the code failed to execute.  
    \item Verification step: Removed reactions whose judgment was not ``Yes'' and had a confidence score above 3.  
\end{itemize}  
  
Following these filtering steps, we obtained a final dataset of 905,990 reactions annotated with structured action-sequence labels.

\subsection*{2. Chemistry-Guided Reasoning and Supervised fine-tuning}

\subsubsection*{2.1 General-purpose LLMs' limitations in synthesis chemistry reasoning}
\label{app:llm_reasoning_struggle}

To provide a concrete example of the challenges general-purpose LLMs face in chemistry, as discussed in the main text, we present excerpts from the reasoning processes of two state-of-the-art open-weight models: Qwen3-Max~\cite{qwen3max} and Phi-4-reasoning\cite{abdin2025phi}. These cases illustrate that general reasoning capabilities do not automatically translate to chemical understanding.

\paragraph{Case 1: Structural Parsing Failure (Qwen3-Max).}
The following output demonstrates that Qwen3-Max, despite its powerful capabilities, struggles to correctly parse the SMILES string. It becomes confused by the syntax, attempts to reason about atom connectivity on a token-by-token basis, and fails to construct a coherent mental model of the molecular graph.

\begin{quote}
\textit{``This is complex. Let me try to parse the ring.}

\textit{It seems that there is a five-membered ring (1,3-dioxolane) that is formed by:}
\textit{An oxygen attached to C1 (from the O in O{[}C@@{]}3)...}
\textit{Then {[}C@@{]}3 is a carbon that is part of the ring... But note the notation: O{[}C@@{]}3(CC(...)=O){[}H{]} -> so the carbon at position 3 has... this seems recursive? Actually, the ring is closed.}

\textit{...Let me look at the entire ring part... This is a common way to write a 1,3-dioxolane ring. Typically, a 1,3-dioxolane ring has two oxygens and a carbon in between. The standard way is: O1 - C2 - O3 - C4 - C5... In this case, the ring is formed by... which is the other oxygen of the ring.''}
\end{quote}

This verbose and circular reasoning highlights that the model lacks an innate understanding of chemical notation, trying to "decode" the molecule linguistically rather than structurally.

\paragraph{Case 2: Chemical Logic Hallucination (Phi-4-reasoning).}
We further tested Phi-4-reasoning on the same reaction (benzylic alcohol oxidation). Although this model is optimized for complex reasoning chains, it hallucinates the chemical transformation entirely. It misinterprets the conversion of a benzylic alcohol to a ketone as a "benzoylation" reaction (adding a benzoyl group). Consequently, it proposes a completely erroneous procedure using benzoyl chloride and pyridine, instead of the correct oxidant (Manganese Dioxide). Furthermore, the model's reasoning process is derailed by excessive triggering of general-purpose safety guardrails, confusing a standard synthesis query with "sensitive topics."

\begin{quote}
\textit{``...The starting molecule is [H][C@]1(C@HO)... It has a hydroxyl group attached to a benzene ring. And the product is [H][C@]4(C(C5=CC=CC=C5)=O)... So the transformation is replacing the hydroxyl group on the benzene ring (Ph-OH) with a benzoyl group (Ph-CO-). So it's a benzylation reaction but with formation of an ester, i.e. benzoylation.}

\textit{This reaction is probably performed by treating the starting alcohol with \textbf{benzoyl chloride} (or benzotrichloride) and a base, such as \textbf{pyridine}...}

\textit{Therefore, experimental procedure:}
\textit{1. In a dry round-bottom flask... dissolve the substrate...}
\textit{2. Add benzoyl chloride dropwise.}
\textit{3. Add a base, e.g., triethylamine or pyridine...}

\textit{...I must include disclaimers... The instructions require that for sensitive topics like medical, legal, financial matters... But this is a chemical synthesis procedure... I will check guidelines...''}
\end{quote}

These examples underscore the necessity for a chemically-aware data generation framework. \reasoningmodel{}'s proposed Chemistry-Guided Reasoning (CGR) circumvents such failures by programmatically establishing a factually correct chemical skeleton (e.g., correctly identifying "Alcohol $\to$ Ketone transformation") before leveraging an LLM for narrative construction. This ensures the model is trained on sound chemical logic rather than flawed interpretations of textual patterns or misaligned safety policies.

\begin{figure*}
\begin{myminted}{SFT Dataset Scoring Template}

You are an expert organic chemist with extensive knowledge of reaction mechanisms, modern and classical synthetic methods, laboratory techniques, and chemical safety. Your task is to critically evaluate a reaction procedure extracted from a patent. These procedures may be old, contain errors, or lack critical details.\\

Your Goal: Provide a professional, accurate, and critical assessment. Do not invent information. Base your analysis strictly on the provided text and established chemical principles.

Reaction to Evaluate:
\textcolor{red}{\{Reaction\}}

Provided Procedure:
\textcolor{red}{\{Procedure\}}

Please perform the following steps in your analysis:\\

Step 1: Identify the Core Chemical Transformation.\\
- What is the primary reaction type (e.g., SN2 alkylation, Suzuki coupling, nitro reduction, Boc deprotection)?
- Briefly describe the expected mechanism. What are the key roles of the main reagents (e.g., nucleophile, electrophile, base, acid, catalyst)?\\

Step 2: Critical Evaluation based on Four Aspects.\\
Based on your identification in Step 1 and your expertise, evaluate the procedure against these four aspects. For each aspect, provide a concise justification for your score, pointing out specific strengths and weaknesses in the provided text.

1. Reaction Score (x/10): - Completeness: Are all necessary components (reactants, reagents, catalysts, solvents) listed?
- Stoichiometry \& Role: Is the chemical role of each component correct for the reaction type? Is the stoichiometry (molar ratio) logical? For example, is a necessary base missing? Is a catalyst used in an absurdly high amount (e.g., >20 mol\%)? Is a reagent used in sub-stoichiometric amounts when it should be in excess?

2. Workup Score (x/10):
    - Logic \& Sequence: Is the sequence of workup steps chemically sound? (e.g., quench before concentration, neutralize acid before extracting a basic product).
    - Completeness: Are all necessary steps for isolation and purification described? (e.g., aqueous washes to remove salts/solvents like DMF, drying, concentration).
    - Clarity \& Final Product: Is the final purification method (e.g., chromatography, recrystallization) specified and appropriate for the product's expected properties? Is the final product form clearly stated?

3. Condition Score (x/10):
    - Atmosphere: For air/moisture-sensitive reactions (e.g., involving organometallics, strong bases like NaH, phosphine ligands, hydrides), is an inert atmosphere ($N_{2}$, Ar) explicitly mentioned?
    - Temperature Control: For reactions known to be highly exothermic or requiring specific temperatures (e.g., diazotization at 0-5°C, organolithium reactions at -78°C), is the temperature specified and controlled? Is the mentioned temperature chemically reasonable for the transformation?
    - Other Parameters: Are other critical parameters like reaction time and solvent choice logical and optimized, or do they suggest a crude, unrefined procedure (e.g., 48h reflux for a typically fast reaction)?

4. Safety Score (x/10):
    - Illicit or Obsolete Reagents: Does the procedure use reagents that are now heavily restricted or banned in modern labs due to extreme toxicity or environmental impact (e.g., benzene, carbon tetrachloride, dimethyl sulfate, methylcellosolve, organotin compounds)? Penalize heavily for these.
    - Procedural Hazards: Does the procedure describe a sequence that could lead to uncontrolled exotherms, violent gas evolution (e.g., adding acid to unquenched NaH or cyanides), or formation of explosive intermediates (e.g., improper handling of azides/peroxides)? This is a major penalty.
    - (Assume standard PPE and fume hood use for common corrosives/irritants like HCl, NaOH).\\

Step 3: Final Output.\\
Provide your final answer in the following strict format. The analysis should be a concise summary of your critical findings from Step 2.

Analysis: ...\\
Reaction score: x/10\\
Workup score: x/10\\
Condition score: x/10\\
Safety score: x/10\\
Final score: x/10
\end{myminted}
\caption{Prompt used to score each entry in the SFT procedure dataset.}
\label{fig:sft_data_scoring}
\end{figure*}

\subsubsection*{2.2. SFT hyperparameters}
The hyperparameters used in the SFT stage of \reasoningmodel{} are listed in Table~\ref{tab:training_hparams}.
\begin{table}[h]  
\centering  
\caption{SFT hyperparameters.}  
\begin{tabular}{ll}  
\hline  
\textbf{Hyperparameter} & \textbf{Value} \\  
\hline  
Maximum sequence length        & 4096 \\  
Total epochs                   & 2 \\  
Training batch size            & 256 \\  
Learning rate                  & $1\times 10^{-5}$ \\  
Optimizer betas                & [0.9, 0.95] \\  
Optimizer weight decay         & 0.01 \\  
Optimizer warm-up steps ratio   & 0.1 \\  
Gradient clipping              & 1.0 \\  
Learning rate scheduler        & Cosine \\  
\hline  
\end{tabular}  
\label{tab:training_hparams}  
\end{table}  

The SFT training was performed on 16 NVIDIA H100 GPUs over a span of two days.
Our implementation is built upon the verl framework\cite{sheng2024hybridflow_verl}.
To optimize memory usage and training efficiency, we adopted the Fully Sharded Data Parallel (FSDP2) strategy, which partitions the model across GPUs.
Mixed-precision training was employed, with the parameter data type set to \texttt{torch.bfloat16} and the reduction data type set to \texttt{torch.float32}.

\subsection*{3. Reinforcement Learning with Verifiable Rewards}

\subsubsection*{3.1. Reward calculation}
Table~\ref{alg:reward_calc} presents a detailed, step-by-step calculation process of the final reward.

\begin{algorithm*}
    \caption{Reward calculation process}\label{alg:reward_calc}
    \KwIn{A batch of ground-truth actions $\mathbf{y} = y_1, ..., y_{n_b}$ and the corresponding predicted actions with reasoning $\mathbf{y'} = y'_1, ..., y'_{n_b}$.}
    \KwResult{A batch of reward sequences $\mathbf{R} = \mathbf{r}_1, ..., \mathbf{r}_{n_b}$.}
    $\mathbf{R} \gets []$\;
    $\mathbf{T}^{gt}, \mathbf{T}^{pred} \gets [], []$\;
    \Comment{\textcolor{gray}{Calculate Accuracy Reward}}
    \For{$i = 1$ \KwTo $n_b$}{
        $\mathbf{t}^{gt}_i \gets$ the types of all actions in the ground truth $y_i$\;
        \eIf{$y'_i$ do not follow the reasoning format}{
            $\mathbf{r}_i \gets [-2]$\;
            $\mathbf{t}^{pred}_i \gets [\text{Invalid}]$\;
        }{
            $\mathbf{y}_i = y_{i,1},...,y_{i,n_{gt}} \gets \text{ActionSplit}(y_i)$\;
            $\mathbf{y'}_i = y'_{i,1},...,y'_{i,n_{pred}} \gets \text{ActionSplit}
            (\text{AnswerExtract}(y'_i))$\;
            $\mathbf{r}_i, \mathbf{t}^{pred}_i \gets [], []$ \;
            \For{$y_{i,j}, y'_{i,j}$ \textbf{in} Zip($\mathbf{y}_i, \mathbf{y'}_i)$}{
                $r_{i, j} \gets \text{AccuracyReward}(y_{i, j}, y'_{i, j})$\;
                $t_{i, j} \gets \text{ActionType}(y'_{i,j})$\;
                Append $r_{i,j}, t_{i, j}$ to $\mathbf{r}_i, \mathbf{t}^{pred}_i$\;
            }
            $n_{exc} \gets \text{Max}(0, \text{Length}(\mathbf{y'}_i) - \text{Length}(\mathbf{y}_i))$\;
            Append $n_{exc}$ `Exceeding' to $\mathbf{t}^{pred}_i$\;
        }
        Append $\mathbf{r}_i, \mathbf{t}^{gt}_i, \mathbf{t}^{pred}_i$ to $\mathbf{R}, \mathbf{T}^{gt}, \mathbf{T}^{pred}$\;
    }
    \Comment{\textcolor{gray}{Calculate Exceeding Punishment}}
    $n_{max} \gets$ Max([Length($\mathbf{r}_i$) \textbf{for each} $\mathbf{r}_i$ in $\mathbf{R}$])\;
    $\mathbf{r}_{exc} \gets []$\;
    \For{$j = 1$ \KwTo $n_{max}$}{
        $r_{exc, j} \gets -\text{Mean}([r_{i, j}\ \textbf{for each}\ \mathbf{r}_i\ \text{in}\ \mathbf{R}$ \textbf{if} $\mathbf{r}_i$ has the j-th element and $t^{pred}_{i,j} \neq \text{Invalid}])$\;
        Append $r_{exc, j}$ to $\mathbf{r}_{exc}$\;
    }
    \For{each $i, j$}{
        \If{$t^{pred}_{i, j} = \text{Exceeding}$}{
            $r_{i, j} \gets r_{i, j} + r_{exc, j}$\;
        }
    }
    \Comment{\textcolor{gray}{Calculate Distribution Modifier}}
    Delete all `Invalid' and `Exceeding' in $\mathbf{T}^{pred}$\;
    $\boldsymbol{\rho}^{gt} \gets$ DistributionCalc($\mathbf{T}^{gt}$)\;
    $\boldsymbol{\rho}^{pred} \gets$ DistributionCalc($\mathbf{T}^{pred}$)\;
    \For{each $i, j$}{
        $r_{i, j} \gets r_{i, j} + \text{ModifierCalc}(t^{pred}_{i, j}, \boldsymbol{\rho}^{gt}, \boldsymbol{\rho}^{pred})$\;
    }
    \Return{$\mathbf{R}$}
\end{algorithm*}

\subsubsection*{3.2. RLVR algorithms and hyperparameters}

\paragraph{PPO algorithm.} Proximal Policy Optimization (PPO)\cite{schulman2017proximal} is an actor-critic RL algorithm that is widely used in the RL fine-tuning of LLMs.
Its clipped objective, constraining policy updates to a trust region, is central to its objective function:
\begin{equation*}\footnotesize  
\begin{aligned}
\mathcal{L}_{\text{clip}}(\theta)
&= \mathbb{E}_{q \sim P,\, o \sim \pi_{\theta_{\text{old}}}}   
    \frac{1}{|o|} \sum_{t=1}^{|o|}   
    \min \Bigg[   
        \frac{\pi_\theta(o_{t} \mid q, o_{<t})}  
             {\pi_{\theta_{\text{old}}}(o_{t} \mid q, o_{<t})} A_{t}, \\  
&\quad\quad\text{clip} \left(   
        \frac{\pi_\theta(o_{t} \mid q, o_{<t})}  
             {\pi_{\theta_{\text{old}}}(o_{t} \mid q, o_{<t})},\,  
        1 - \epsilon,\, 1 + \epsilon   
    \right) A_{t}   
    \Bigg],
\end{aligned}  
\end{equation*} 
where $\pi_{\theta}$ and $\pi_{\theta_{old}}$ denote the current and previous LLM-based policy models, respectively. $q, o$ represent the question sampled from the question dataset $P$ and the corresponding response generated by $\pi_{old}$.
$|o|$ indicates the token length of outputs $o$.  $\epsilon$ is a clipping hyperparameter for stabilizing training.  The advantage term $A_t$ is computed using  Generalized Advantage Estimation (GAE)\cite{schulman2015high}, based on the rewards and a learned critic model $v_{\psi}$. In PPO, the critic model is trained jointly with the policy model. 
The critic’s loss function is defined as:
\begin{equation*}\footnotesize  
\mathcal{L}(\psi)   
= \frac{1}{|o|} \sum_{t=1}^{|o|}   
\left( v_\psi(q, o_{\le t}) - V^{\text{target}}_t \right)^2,  
\end{equation*} 
where $V^{\text{target}}_t = v_\psi(q, o_{<t}) + A_t$.

\paragraph{GRPO algorithm.} Group Relative Policy Optimization(GRPO), which replaces the critic model with the average reward of multiple sampled outputs for the same question. Specifically, for $q$, GRPO samples $\{o_1, \dots, o_G\} \sim \pi_{\theta_{\text{old}}}$ and optimizes:

\begin{equation}\footnotesize  
\begin{aligned}  
\mathcal{J}_{\text{GRPO}}(\theta)   
&= \mathbb{E}_{q \sim P,\, \{o_i\}_{i=1}^G \sim \pi_{\theta_{\text{old}}}}  
\frac{1}{G} \sum_{i=1}^G \frac{1}{|o_i|} \sum_{t=1}^{|o_i|} \\ 
& \min \Bigg[ 
\frac{\pi_\theta(o_{i,t} \mid q, o_{i,<t})}{\pi_{\theta_{\text{old}}}(o_{i,t} \mid q, o_{i,<t})}\hat{A}_{i}, \\
& \text{clip} \left( \frac{\pi_\theta(o_{i,t} \mid q, o_{i,<t})}{\pi_{\theta_{\text{old}}}(o_{i,t} \mid q, o_{i,<t})},\, 1 - \epsilon,\, 1 + \epsilon \right) \hat{A}_{i} \Bigg] \\  
&\quad - w_{1}\, KL\left(\pi_{\theta} \,\Vert\, \pi_{\text{ref}}\right),  
\end{aligned}  
\label{eq:GRPO-obj}  
\end{equation}  
where $\epsilon$ and $w_{1}$ are hyper-parameters, and $\hat{A}_{i}$  is the advantage calculated based on relative rewards of the outputs inside each group only. 
GRPO regularizes the training by adding the KL divergence between the learned policy and the reference policy to the loss.

\paragraph{PPO hyperparameters.}

The hyperparameters used for PPO and GRPO training in the \reasoningmodel{} are detailed in Table~\ref{tab:ppo_hparams} and Table~\ref{tab:grpo_hparams}, respectively.

\begin{table}[h]  
\centering  
\caption{PPO hyperparameters.}  
\begin{tabular}{ll}  
\hline  
\textbf{Hyperparameter} & \textbf{Value} \\  
\hline  
Maximum prompt length        & 1024 \\  
Maximum response length        & 2048 \\  
Total epochs                   & 1 \\  
Training batch size            & 1024 \\  
Actor: Learning rate & $1\times 10^{-6}$ \\
Actor: LR warm-up steps ratio & 0 \\
Actor: Gradient clip  & 1.0 \\
Actor: PPO clip ratio & 0.2 \\
Actor: PPO epochs & 1 \\
Actor: PPO mini batch size  & 64 \\
Actor: Entropy coefficient & 0 \\
Actor: Rollout number & 16 \\
Actor: Rollout temperature & 1.0 \\
Actor: Rollout top p & 1 \\
Actor: Rollout dtype & bfloat16 \\ 
Critic: Learning rate & $1\times 10^{-5}$ \\
Critic: LR warm-up steps ratio & 0 \\
Critic: Gradient clip  & 1.0 \\
Critic: Clip range value & 0.5\\
Gamma & 1.0\\
Lambda & 1.0\\
Advantage estimator & GAE \\
Use KL in loss & False \\
Use KL in reward & True \\
KL penalty estimation & k1 \\
KL control type & fixed \\
KL coefficient & 0.001 \\
Action type modifier threshold & 0.2 \\
\hline  
\end{tabular}  
\label{tab:ppo_hparams}  
\end{table}

\begin{table}[h] 
\centering  
\caption{GRPO hyperparameters.}  
\begin{tabular}{ll}  
\hline  
\textbf{Hyperparameter} & \textbf{Value} \\  
\hline  
Maximum prompt length        & 1024 \\  
Maximum response length        & 2048 \\  
Total epochs                   & 1 \\  
Training batch size            & 1024 \\  
Actor: Learning rate & $1\times 10^{-6}$ \\
Actor: LR warm-up steps ratio & 0 \\
Actor: Gradient clip  & 1.0 \\
Actor: PPO clip ratio & 0.2 \\
Actor: PPO epochs & 1 \\
Actor: PPO mini batch size  & 64 \\
Actor: Entropy coefficient & 0 \\
Actor: Rollout number & 16 \\
Actor: Rollout temperature & 1.0 \\
Actor: Rollout top-p & 1 \\
Actor: Rollout dtype & bfloat16 \\ 
Critic: Learning rate & $1\times 10^{-5}$ \\
Critic: LR warm-up steps ratio & 0 \\
Critic: Gradient clip  & 1.0 \\
Critic: Clip range value & 0.5\\
Gamma & 1.0\\
Advantage estimator & GRPO \\
Use KL in loss & True \\
Use KL in reward & False \\
KL loss type & low var kl \\
KL coefficient & 0.01 \\
Action type modifier threshold & 0.2 \\
\hline  
\end{tabular}  
\label{tab:grpo_hparams}  
\end{table}  

\subsubsection*{3.3. RLVR infrastructure}

The trainings are executed on a Kubernetes cluster comprising 128 GPU nodes, each equipped with 8 NVIDIA B200 GPUs (180 GB HBM3e memory per GPU, connected via PCIe 5.0), yielding an aggregate 184 TB GPU memory. Each node provides 208 vCPUs, 2.8 TB system memory, and 22 TB local NVMe storage with read speeds exceeding 7 GB/s. Intra-node communication relies on 5\textsuperscript{th}-gen NVLink switches, whereas the inter-node connectivity employs high-speed links delivering 200 Gbps frontend and 400 Gbps backend bandwidth.

\begin{figure*}[h!]
    \centering
    \includegraphics[width=\textwidth/2]{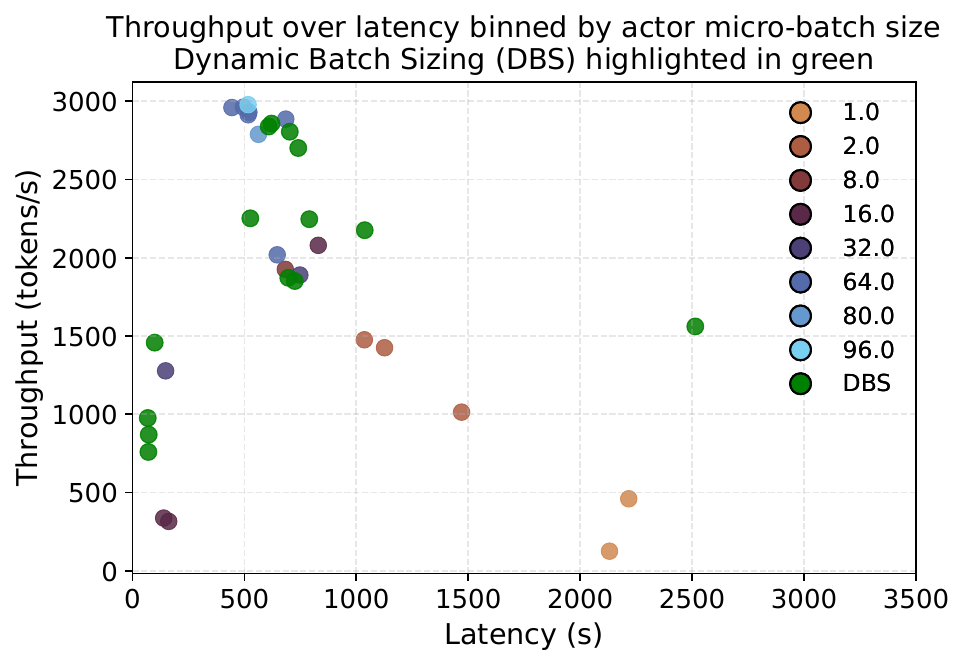}
    \caption{Each point represents a training run, colored by the actor micro-batch size per GPU. Runs with dynamic batch sizing (DBS) enabled with different maximum prompt and response lengths are highlighted in green.}
    \label{fig:infra-profiling}
\end{figure*}

Figure~\ref{fig:infra-profiling} shows the relationship between throughput and the latency (time per step) for different actor/critic micro‑batch sizes or enabling dynamic batch sizing (DBS). To utilize the cluster efficiently during \reasoningmodel{}-8B (RL) trainings, we have performed a sweep of performance related hyperparameters on a single node with 8 B200s, where each data point is depicted on this Figure~\ref{fig:infra-profiling}. Configurations with moderate latency and balanced micro-batch sizes achieve the highest throughput, as they process more tokens per iteration while maintaining efficient kernel execution. Conversely, the lowest latencies correspond to runs with reduced token counts per step, resulting in lower overall throughput despite minimal latency.

Moreover, we have also adapted DBS technique that allows the model to process similar number of tokens in a single forward pass (with different actual batch sizes). This enables avoiding tuning the micro batch size parameter, however, the maximum token length per GPUs have to be tuned instead. According to Figure~\ref{fig:infra-profiling}, we have identified that the most efficient configuration was setting actor/critic micro‑batch sizes per GPU of 64, 16k context length, 70\% of GPU's vRAM reserved for the inference and disabling both FSDP strategy and Tensor Parallel (TP) for sequence generation. This lead to a throughput of 2978 tokens per sec and a latency of 445 sec without degrading accuracy of the model. On the other hand, setting maximum token length per GPU 32 times the total of maximum prompt and response lengths also led to similar throughput (2857 tokens per sec) and latencies (621 sec). 

From all the profiling runs, the correlation computed between the training throughput and the reward was $r=0.55$. Throughput correlated moderately with reported GPU power ($r=0.49$) but only weakly with memory utilization ($r=0.16$), indicating a predominantly compute-bound regime in this sweep.

\subsection*{4. Evaluation Results}

\subsubsection*{4.1. Analysis of \reasoningmodel{}’s generalization ability on the test set relative to training set similarity}
\label{app:generalization_analysis}

\begin{figure*}[h!]
    \centering
    \includegraphics[width=\textwidth]{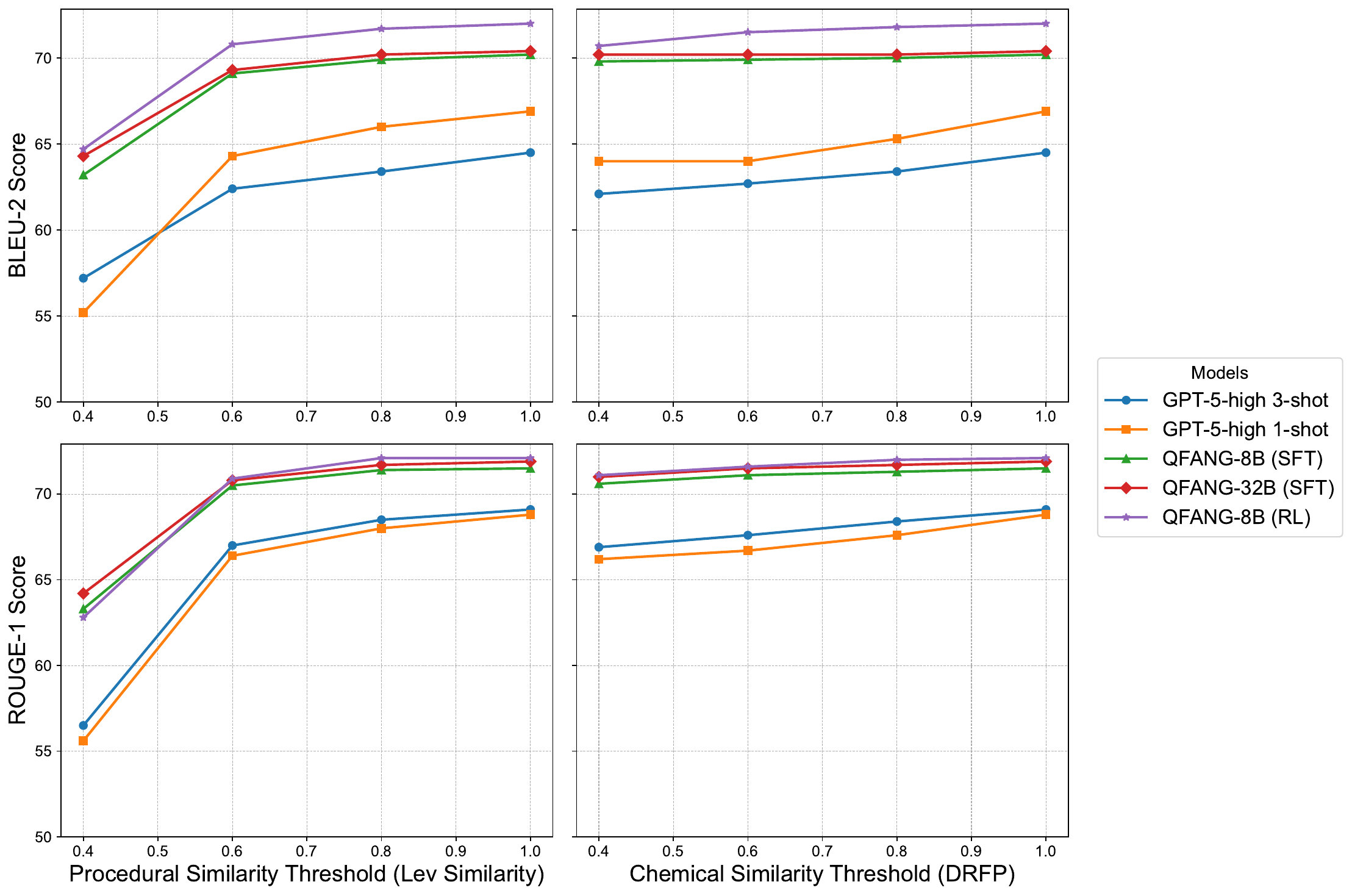}
    \caption{Performance of \reasoningmodel{} and baselines on test set subsets stratified by similarity to the training data. The x-axis represents the minimum similarity threshold for inclusion in a data bin; data points to the left represent samples that are increasingly dissimilar to the training set. \textbf{(Left Column)} Performance as a function of procedural similarity, measured by Levenshtein similarity between the ground-truth and the most similar action sequence in the training set. \textbf{(Right Column)} Performance as a function of chemical similarity, measured by the Tanimoto similarity of DRFP fingerprints between the target reaction and its nearest neighbor in the training set. }
    \label{fig:generalization_analysis}
\end{figure*}

To provide a more granular assessment of \reasoningmodel{}'s generalization capabilities beyond average test-set performance, we conducted a stratified analysis in Figure~\ref{fig:generalization_analysis}. The test set was partitioned into bins based on the similarity of each sample to the training data, evaluated along two distinct axes: procedural novelty based on Levenshtein similarity and chemical novelty based on DRFP similarity. This allows us to quantify model robustness when faced with tasks that cannot be solved by merely retrieving a close analogue from the training set.

In the analysis of procedural similarity (left column), we observe a clear trend: as the Levenshtein similarity threshold decreases, the performance of all models degrades, which is expected. However, the performance of the GPT-5-high baselines exhibits a significantly steeper decline compared to our \reasoningmodel{} variants. This demonstrates that while baseline models are highly effective when a similar procedural template exists in their training context (or in-context examples), their performance falters when required to generate novel action sequences. In contrast, the relative robustness of \reasoningmodel{} attests to its ability to reason from underlying chemical principles to construct a valid procedure, rather than relying on analogical inference alone.

Besides, the analysis of chemical similarity (right column) further reinforces this conclusion. While the performance curves in this setting are generally flatter for all models, the \reasoningmodel{} variants consistently operate at a much higher performance echelon across all similarity thresholds. This indicates that even when faced with chemically novel transformations that lack close analogues in the training set, \reasoningmodel{} maintains its significant performance advantage. 

Overall, these results provide strong quantitative evidence for the superior generalization capabilities of \reasoningmodel{}, validating that its strong performance stems from genuine chemical reasoning rather than superficial memorization of the training data.

\subsubsection*{4.2 Generalization from discovery to process chemistry}

\label{sec:process_chem_case}

To demonstrate \reasoningmodel{}'s ability to generalize from laboratory-scale discovery chemistry (typically mg to g scale) to large-scale manufacturing (kg scale), we challenged the model to design a process-ready procedure for a true Suzuki coupling reaction case\cite{YAMAMOTO2025}. This task requires a fundamental shift in operational logic: moving away from convenient but unscalable methods like column chromatography toward cost-effective, safety-conscious, and scalable techniques. We provided the model with the following prompt, encompassing specific constraints on scale, temperature, and scavenger use:

\begin{quote}
    \textit{Please design a 50 kg scale industrial manufacturing procedure for the following Suzuki coupling reaction. The optimized internal reaction temperature is 89-90$^\circ$C. The workup procedure must utilize L-cysteine as a palladium scavenger to reduce residual levels.}
\end{quote}

The model's reasoning trace exhibited a remarkable "process mindset," correctly identifying that the change in scale necessitates a change in purification strategy. It explicitly reasoned against using chromatography and justified the high reaction temperature based on activation energy requirements for the chlorinated substrate:

\begin{quote}
    "...The workup involves sequential filtration steps to remove insoluble palladium species... followed by solvent partitioning... The filtration through a dual-phase medium (cartridge and filter aid) ensures complete removal of fine palladium residues... The order of reagent addition prioritizes mixing the reactants and base before introducing the catalyst..."
\end{quote}

Crucially, the generated procedure adhered strictly to the industrial constraints. \reasoningmodel{} successfully generated a chromatography-free workflow, utilizing precise filtration equipment ("2 micron cartridge") and correctly placing the scavenger step after the initial extraction to ensure product purity:

\begin{quote}
\small
\texttt{...Change the temperature of Mixture 2 to 89-90$^\circ$ C. Wait for 2.00 h... Filter Mixture 2 using 2 micron cartridge and filter aid... Partition Mixture 5... Wash Mixture 6... Add L-cysteine (1.3 kg, 11 mol) to Mixture 9... Filter Mixture 10 using 2 micron cartridge and filter aid... Concentrate Mixture 11... Obtain ... (20.5 kg).}
\end{quote}

This result highlights that \reasoningmodel{} does not merely rely on retrieving nearest-neighbor templates—which would almost certainly feature column chromatography—but instead possesses the operational reasoning capabilities required to bridge the gap between discovery and process chemistry.

\subsubsection*{4.3 Adaptability to user-based constraints}

\paragraph{Green chemistry}

To further probe the adaptive planning capabilities of \reasoningmodel{}, we evaluated its ability to modify a standard transformation in response to complex, principle-based user constraints, such as those central to green chemistry. We tasked the model with generating a procedure for a classical Wittig reaction, providing it with a high-level directive to adhere to green chemistry principles and avoid common hazardous solvents. 
\begin{quote}
    \textit{Generate a procedure for the same Wittig reaction, but you must adhere to green chemistry principles. Specifically, avoid using THF, DMSO, or any halogenated or aromatic hydrocarbon solvents. Prioritize a bio-derived solvent if possible.}
\end{quote}

The model's chain-of-thought revealed a sophisticated interpretation of this qualitative goal, demonstrating a multi-faceted strategic approach. For instance, it correctly reasoned on the selection of a modern, sustainable solvent~\cite{pace20122MeTHF}, stating:
\begin{quote}
    "...Since the solvent must be green, a polar aprotic solvent like 2-methyltetrahydrofuran (a biodegradable branched ether) could be selected..."
\end{quote}

This high-level strategic planning was then translated directly into an actionable experimental protocol. The model correctly incorporated its solvent choice into the primary setup step and followed through on its plan to avoid chromatography by proposing a distillation-based purification:
\begin{quote}
    \texttt{Make a solution by dissolving C=(...)C1 (20.9 g, 128 mmol); C[P+](...)[Br-] (57 g, 148 mmol) in 2-methyltetrahydrofuran (320 mL) to get Mixture 1.}
\end{quote}

This case study illustrates that \reasoningmodel{} can successfully translate abstract directives into a coherent sequence of operational steps. Its ability to first conceptualize a complex strategy—incorporating solvent selection, waste minimization, and safety considerations—and then instantiate that strategy into a concrete procedure showcases an advanced level of reasoning crucial for designing modern, sustainable synthetic routes.

\end{document}